\documentclass[10pt,journal,compsoc]{IEEEtran}

\usepackage{amsmath}
\usepackage{diagbox}
\usepackage{graphics}
\usepackage{epsfig}
\usepackage{threeparttable}
\usepackage{xspace}
\usepackage{siunitx}
\usepackage{graphicx}
\usepackage{amssymb}
\usepackage{threeparttable}
\usepackage{color}
\usepackage[normalem]{ulem}
\usepackage{multirow}
\usepackage{float}
\usepackage{amsfonts}
\usepackage{bm}
\usepackage{array}
\usepackage[table]{xcolor}
\usepackage{colortbl}
\usepackage{diagbox}
\usepackage{rotating}
\usepackage{booktabs}
\usepackage{overpic}
\usepackage{enumitem}
\usepackage{colortbl}
\usepackage{algorithm}
\usepackage{algorithmic}
\usepackage{cite}
\usepackage{dblfloatfix}

\usepackage[pagebackref=false,breaklinks=true,colorlinks,bookmarks=false]{hyperref}

\definecolor{mygray}{gray}{.85}
\definecolor{mygray1}{gray}{.7}
\definecolor{mygray2}{gray}{.93}

\newcommand{\tabincell}[2]{\begin{tabular}{@{}#1@{}}#2\end{tabular}}

\makeatletter
\newcommand{\thickhline}{%
	\noalign {\ifnum 0=`}\fi \hrule height 1pt
	\futurelet \reserved@a \@xhline
}
\makeatother

\makeatletter
\DeclareRobustCommand\onedot{\futurelet\@let@token\@onedot}
\def\@onedot{\ifx\@let@token.\else.\null\fi\xspace}

\def\eg{\emph{e.g}\onedot} 
\def\ie{\emph{i.e}\onedot} 
 
\def\etc{\emph{etc}\onedot} 
 
\def\etal{\emph{et al}\onedot}

\makeatother

\newcommand{\red}[1]{{\textcolor{black}{{#1}}}}

\begin{document}
	\title{A Survey on Deep Learning Technique for\\ Video Segmentation}

	\author{Tianfei Zhou, Fatih Porikli,~\IEEEmembership{Fellow~IEEE}, David J. Crandall,~\IEEEmembership{Senior~Member~IEEE},\\ Luc Van Gool, Wenguan~Wang,~\IEEEmembership{Senior~Member~IEEE}
		\IEEEcompsocitemizethanks{
			\IEEEcompsocthanksitem T. Zhou and L. Van Gool are with ETH Zurich. (Email: ztfei.debug@gmail.com, vangool@vision.ee.ethz.ch)
			\IEEEcompsocthanksitem F. Porikli is with the School of Computer Science, Australian National University. (Email: fatih.porikli@anu.edu.au)
			\IEEEcompsocthanksitem D. Crandall is with the Luddy School of Informatics, Computing, and Engineering, Indiana University.
			(Email: djcran@indiana.edu)
			\IEEEcompsocthanksitem W. Wang is with ReLER Lab, Australian Artificial Intelligence Institute, University of Technology Sydney (Email: wenguanwang.ai@gmail.com)
			\IEEEcompsocthanksitem Corresponding author: \textit{Wenguan Wang}
			
			%
		}
	}

\markboth{IEEE TRANSACTIONS ON PATTERN ANALYSIS AND MACHINE INTELLIGENCE}%
{Shell \MakeLowercase{\textit{et al.}}: Bare Demo of IEEEtran.cls for Journals}

\IEEEtitleabstractindextext{
\begin{abstract}
	Video segmentation---partitioning video frames into multiple segments or objects---plays a critical role in a broad range of  practical applications, {from enhancing visual effects in movie, to understanding scenes in autonomous driving, to creating virtual background in video conferencing.} Recently, with the renaissance of connectionism in computer vision, there has been {an influx of deep learning based approaches for video segmentation that have delivered compelling performance.} In this survey, {we comprehensively review two basic lines of research --- generic object segmentation (of unknown categories) in videos, and video semantic segmentation --- by introducing their respective task settings,} background concepts, perceived need, development history, and main challenges. We also offer a detailed overview of representative literature on both methods and datasets. We further benchmark the reviewed methods on several well-known datasets. Finally, we point out open issues in this field, and suggest opportunities for further research. {We also provide a public website to continuously track developments in this fast advancing field: \url{https://github.com/tfzhou/VS-Survey}.}
\end{abstract}
}

\maketitle
\IEEEdisplaynontitleabstractindextext
\IEEEpeerreviewmaketitle

\IEEEraisesectionheading{\section{Introduction}\label{sec:introduction}}
\label{sec:intro}

\IEEEPARstart{V}{ideo} segmentation --- identifying the key
objects with some specific properties or semantics in a
video scene --- is a fundamental and challenging problem in computer vision, with
numerous potential applications including autonomous driving, robotics, automated surveillance, social media,  augmented reality, movie production, and video conferencing.

The problem has been addressed  using various traditional computer vision and machine learning techniques, including hand-crafted features (\eg, histogram statistics, optical flow, \etc), heuristic prior knowledge (\eg, visual attention mechanism$_{\!}$~\cite{DBLP:conf/cvpr/WangSP15}, motion boundaries$_{\!}$~\cite{DBLP:conf/iccv/PapazoglouF13}, \etc), low/mid-level visual representations (\eg, super-voxel$_{\!}$~\cite{xu2012evaluation}, trajectory$_{\!}$~\cite{Brox2010}, object proposal$_{\!}$~\cite{lee2011key}, \etc), and classical machine learning models (\eg, clustering$_{\!}$~\cite{yu2015efficient}, graph models$_{\!}$~\cite{grundmann2010}, random walks$_{\!}$~\cite{shankar2015}, support vector machines$_{\!}$~\cite{perazzi2015fully}, random decision forests$_{\!}$~\cite{badrinarayanan2013semi}, markov random fields$_{\!}$~\cite{jang2016streaming}, conditional random fields$_{\!}$~\cite{liu2015multiclass}, \etc).
Recently, deep neural networks, and Fully Convolutional Networks (FCNs)$_{\!}$~\cite{long2015fully} in particular, have led to remarkable advances in video segmentation. These deep learning-based video segmentation algorithms are significantly more accurate (and sometimes even more efficient) than traditional approaches.

\begin{figure}[!htb]
	\vspace{-26pt}
	\centering
	\includegraphics[width=0.99\linewidth]{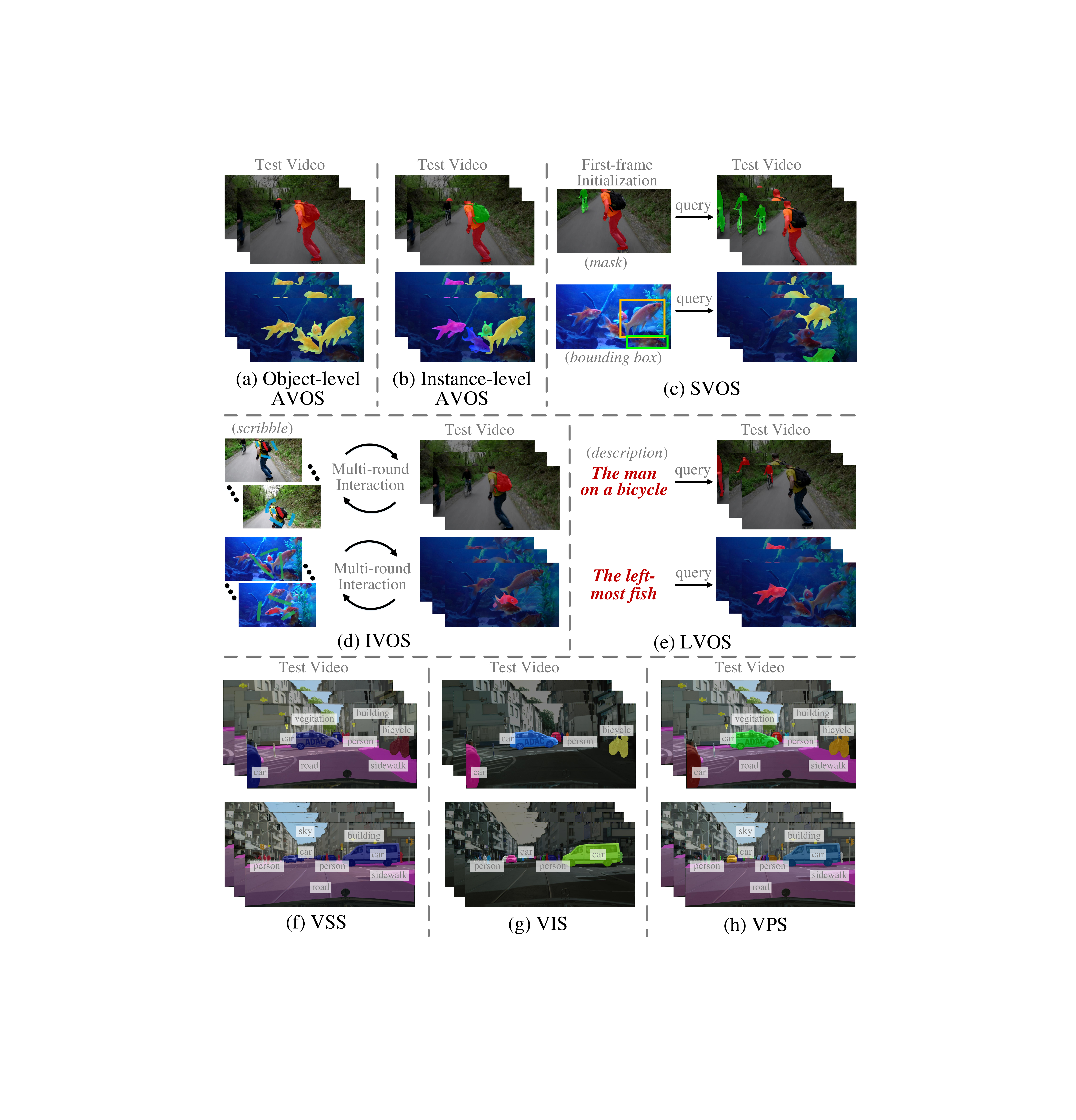}
	\vspace{-10pt}
	\caption{Video segmentation tasks reviewed in this survey: (a) object-level automatic video object segmentation (object-level AVOS), (b) instance-level automatic video object segmentation (instance-level AVOS), (c) semi-automatic video object segmentation (SVOS), (d) interactive video object segmentation (IVOS), (e) language-guided video object segmentation (LVOS),  (f) video semantic segmentation (VSS), (g) video instance segmentation (VIS), and (h) video panoptic segmentation (VPS).
	}
	\vspace{-12pt}
	\label{fig:overview}
\end{figure}

\begin{figure*}[!htb]
	\centering
	\includegraphics[width=0.99\linewidth]{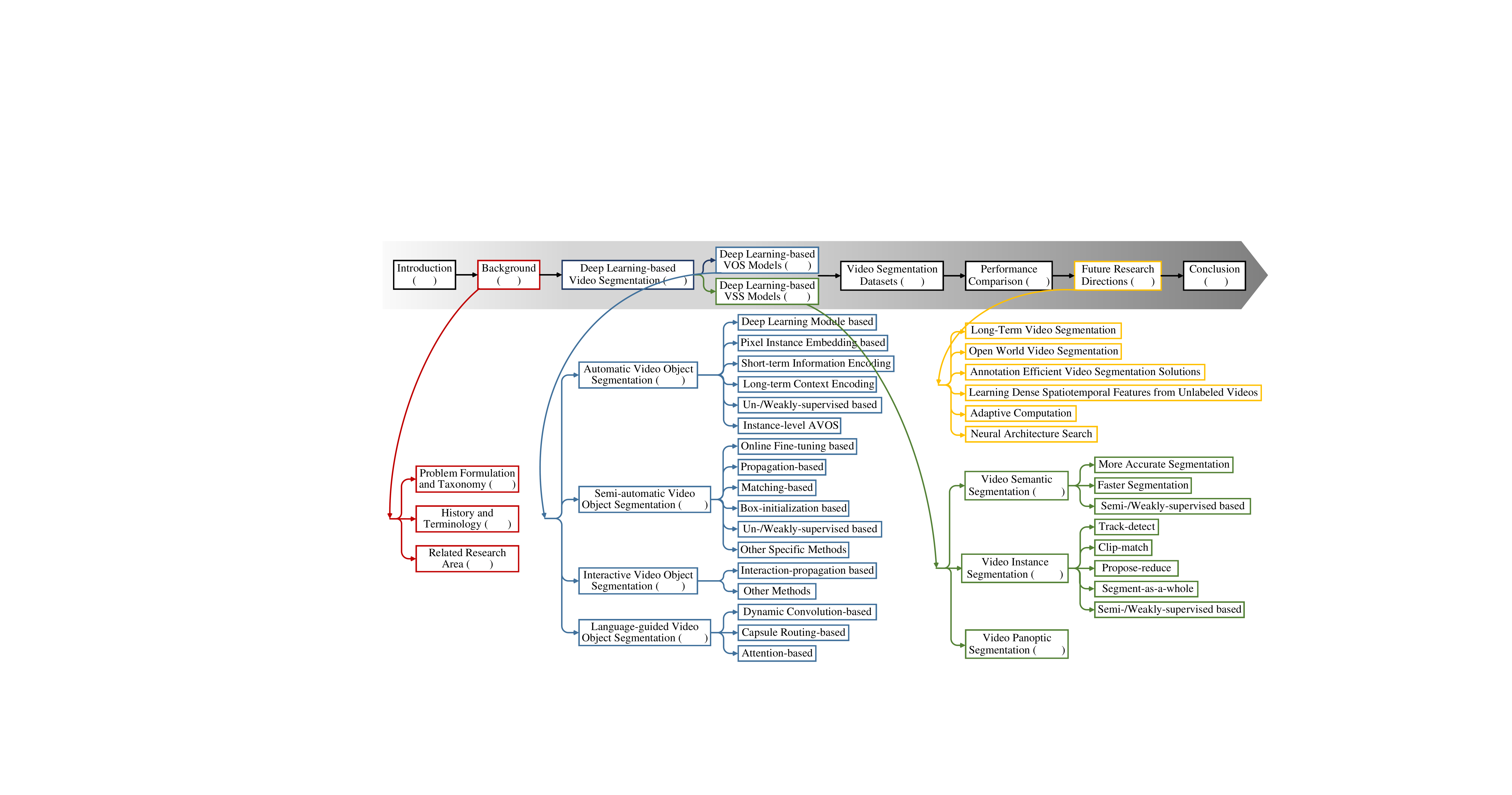}
	\put(-492,219){\scriptsize \S\ref{sec:introduction}}
	\put(-443.5,219){\scriptsize \S\ref{sec:2}}
	\put(-348,219){\scriptsize \S\ref{sec:3}}
	\put(-279.5,228){\scriptsize \S\ref{sec:3.1}}
	\put(-280,210){\scriptsize \S\ref{sec:3.2}}
	\put(-210.5,218.5){\scriptsize \S\ref{sec:4}}
	\put(-138,218.5){\scriptsize \S\ref{sec:5}}
	\put(-77.5,218.5){\scriptsize \S\ref{sec:6}}
	\put(-35,218.5){\scriptsize \S\ref{sec:7}}
	\put(-448.5,102){\tiny \S\ref{sec:2.1}}
	\put(-451.5,79){\tiny \S\ref{sec:2.2}}
	\put(-461,56){\tiny \S\ref{sec:2.3}}
	\put(-353,162){\tiny \S\ref{sec:3.1.1}}
	\put(-340.5,90){\tiny \S\ref{sec:3.1.2}}
	\put(-353,43){\tiny \S\ref{sec:3.1.3}}
	\put(-340.5,13){\tiny \S\ref{sec:3.1.4}}
	\put(-134.5,98){\tiny \S\ref{sec:vss}}
	\put(-135.5,50){\tiny \S\ref{sec:vis}}
	\put(-134.5,6){\tiny \S\ref{sec:vps}}
	\vspace{-12pt}
	\caption{{Overview of this survey.}}
	\vspace{-10pt}
	\label{fig:overview2}
\end{figure*}

With the rapid advance of this field, there is a huge body of new literature being produced. However, {most existing surveys predate the modern deep learning era$_{\!}$~\cite{thounaojam2014survey,zhang2006overview},} and often take a narrow view, such as focusing only on video foreground/background segmentation$_{\!}$~\cite{yao2020video,perazzi2016benchmark}. 
In this paper, we offer
a  state-of-the-art review that addresses the
wide area$_{\!}$ of$_{\!}$ video$_{\!}$ segmentation,$_{\!}$ especially$_{\!}$ to$_{\!}$ help$_{\!}$ new$_{\!}$ researchers$_{\!}$
enter this rapidly-developing field. We systematically introduce
recent advances in video segmentation, spanning from task formulation to taxonomy, from algorithms to datasets, and from unsolved issues to future research directions. We cover crucial aspects including task categories (\ie, foreground/background separation \textit{vs} semantic segmentation), {inference modes} (\ie, automatic, semi-automatic, and interactive), and learning paradigms (\ie, supervised, unsupervised, and weakly supervised), and we try to clarify  terminology (\eg, background subtraction, motion segmentation, \etc). \red{We hope that this survey helps accelerate progress in this field.}

This survey mainly focuses on recent progress in two major branches of video segmentation, namely video object segmentation (Fig.~\ref{fig:overview}(a-e)) and video semantic segmentation  (Fig.$_{\!}$~\ref{fig:overview}(f-h)),$_{\!}$ which$_{\!}$ are$_{\!}$ further$_{\!}$ divided$_{\!}$ into$_{\!}$ eight$_{\!}$ sub-fields. {Even
	after restricting our focus to deep learning-based video segmentation, there are still hundreds of papers in this fast-growing field. We select influential work published in prestigious journals and conferences. We also  include some non-deep learning video segmentation models and relevant literature in other areas, \eg, visual tracking, to give necessary background.} \red{Moreover, in order to promote the development of this field, we provide an accompanying webpage which catalogs algorithms and datasets addressing
	video segmentation: \url{https://github.com/tfzhou/VS-Survey}.}

{Fig.~\ref{fig:overview2} shows the structure of this survey.}  Section \S\ref{sec:2} gives some brief background on taxonomy, terminology, study history, and related research areas. We review representative papers on deep learning algorithms and video segmentation datasets in \S\ref{sec:3} and \S\ref{sec:4}, respectively.  Section \S\ref{sec:5} conducts performance evaluation and analysis, {while \S\ref{sec:6} raises open questions and directions.} Finally, we make concluding remarks in \S\ref{sec:7}.

\section{Background}\label{sec:2}
In this section, we first formalize the task, categorize research directions, and discuss key challenges and driving factors in \S\ref{sec:2.1}. Then, \S\ref{sec:2.2}~offers a brief historical background covering early work and foundations, and \S\ref{sec:2.3} establishes linkages with  relevant fields.
		
\vspace{-2pt}
\subsection{Problem Formulation and Taxonomy}\label{sec:2.1}
Formally, let \bm{$\mathcal{X}$} and \bm{$\mathcal{Y}$} denote the input space and output segmentation space, respectively. Deep learning-based video segmentation solutions generally seek to learn an \textit{ideal} video-to-segment mapping $f^{*\!}:\bm{\mathcal{X}}\mapsto\bm{\mathcal{Y}}$.

\vspace{-2pt}
\subsubsection{Video Segmentation Category}\label{sec:2.1.1}
According to how the output space \bm{$\mathcal{Y}$} is defined, video segmentation can be broadly categorized into two classes: video object (foreground/background) segmentation, and video semantic segmentation.

\noindent$\bullet$~\textbf{Video Foreground/Background Segmentation (Video Object Segmentation, VOS)}. VOS is the classic video segmentation setting and refers to segmenting dominant objects (of unknown categories). In this case, $\bm{\mathcal{Y}}$ is a binary, foreground/background segmentation space. VOS is typically used in video analysis and editing application scenarios, such as object removal in movie editing, content-based video coding, and virtual background creation in video conferencing. It typically is not concerned with the exact semantic categories of the segmented objects.

\noindent$\bullet$~\textbf{Video Semantic Segmentation (VSS)}. As a direct extension of image semantic segmentation to the spatio-temporal domain, VSS aims to extract objects within predefined semantic categories (\eg, car, building, pedestrian, road) from videos. Thus, $\bm{\mathcal{Y}}$ corresponds to a multi-class, semantic parsing space. VSS serves as a perception foundation for many application fields, such as robot sensing, human-machine interaction, and autonomous driving, which require high-level understanding of the physical environment.

\noindent\textbf{Remark}. VOS and VSS share some common challenges, such as fast motion and object occlusion. However, due to differences in application scenarios, many challenges are different. For instance, {VOS often focuses on  human created media, which often have large camera motion, deformation, and appearance changes. VSS instead often focuses on applications like autonomous driving, which requires a good trade off between accuracy and latency, accurate detection of small objects, model parallelization, and cross-domain generalization ability.}

\subsubsection{{Inference Modes} for Video Segmentation}\label{sec:2.1.2}
VOS methods can be further classified into three types: automatic, semi-automatic, and interactive, according to how much human intervention is involved during inference.

\noindent$\bullet$~\textbf{Automatic Video Object Segmentation (AVOS)}. AVOS, or \textit{unsupervised video segmentation} or \textit{zero-shot video segmentation}, performs VOS in an automatic manner, without any manual initialization  (Fig.~\ref{fig:overview}(a-b)). The input space $\bm{\mathcal{X}}$ refers to the video domain $\bm{\mathcal{V}}$ only. AVOS is suitable for video analysis but not for video editing that requires segmenting arbitrary objects or their parts flexibly; a typical application is virtual background creation in video conferencing. 

\noindent$\bullet$~\textbf{Semi-automatic Video Object Segmentation (SVOS)}. SVOS, also known as \textit{semi-supervised video segmentation} or \textit{one-shot video segmentation}~\cite{DBLP:conf/cvpr/CaellesMPLCG17}, involves limited human inspection (typically provided in the first frame) to specify the desired objects  (Fig.~\ref{fig:overview}(c)). For SVOS,  $\bm{\mathcal{X}}_{\!}\!=_{\!}\!\bm{\mathcal{V}}\!\times\!\bm{\mathcal{M}}$, where $\bm{\mathcal{V}}_{\!}$ indicates the video space and~$\bm{\mathcal{M}}$ refers to human input. Typically the human input is an object mask in the  first video frame, in which case SVOS is also called \textit{pixel-wise} \textit{tracking} or \textit{mask propagation}.
{Other forms of human input include bounding boxes and scribbles~\cite{shankar2015}.} From this perspective, \textbf{language-guided video object segmentation (LVOS)} is a sub-branch of SVOS, in which the human input is given as linguistic descriptions about the desired objects  (Fig.~\ref{fig:overview}(e)).
Compared to AVOS, {SVOS is more flexible in defining target objects, but requires human input.} SVOS is typically applied in a user-friendly setting (without specialized equipment), such as video content creation in mobile phones. One of the core challenges in SVOS is how to fully utilize target information from limited human intervention.

\noindent$\bullet$~\textbf{Interactive Video Object Segmentation (IVOS)}. SVOS models are designed to operate automatically once the target has been identified, while systems for IVOS incorporate user guidance throughout the analysis process  (Fig.~\ref{fig:overview}(d)). IVOS can obtain high-quality segments and works well for computer-generated imagery and video post-production, {where tedious human supervision is possible. IVOS is also studied in the graphics community as \textit{video cutout}.} The input space $\bm{\mathcal{X}}$ for IVOS is $\bm{\mathcal{V}}\!\times\!\bm{\mathcal{S}}$, where $\bm{\mathcal{S}}$ typically refers to human scribbling.  Key challenges include: 1) allowing users to easily specify segmentation constraints; 2) incorporating human specified constraints into the segmentation algorithm; and 3) giving quick response to the constraints.

In contrast to VOS,
VSS methods typically work in an automatic mode (Fig.~\ref{fig:overview}(f-h)), \ie, $\bm{\mathcal{X}}\!\equiv\!\bm{\mathcal{V}}$. Only a few early methods  address the semi-automatic setting, called \textit{label propagation}~\cite{badrinarayanan2010label}.

\noindent\textbf{Remark}. The terms ``unsupervised'' and ``semi-supervised'' are conventionally used in VOS to specify the amount of human interaction involved during inference. But they are easily confused with ``unsupervised$_{\!}$ learning''$_{\!}$ and$_{\!}$ ``semi-supervised$_{\!}$ learning.''$_{\!}$ We$_{\!}$ urge$_{\!}$ the$_{\!}$ community$_{\!}$ to$_{\!}$ replace$_{\!}$ these$_{\!}$  ambiguous$_{\!}$ terms with ``automatic'' and ``semi-automatic.''

\vspace{-3pt}
\subsubsection{Learning Paradigms for Video Segmentation}\label{sec:2.1.3}
Deep learning-based video segmentation models can be grouped into three categories
according to the learning strategy they use to approximate $f^{*}$:
supervised, unsupervised, and weakly supervised.

\noindent$\bullet$~\textbf{Supervised Learning Methods.} Modern video segmentation models are typically learned in a fully supervised manner, requiring $N$ input training samples and their desired outputs $y_n\!\!:=\!f^{*\!}(x_n)$, where $\{(x_n, y_n)\}_{n\!}\!\subset$\bm{$\mathcal{X}$}$\times$\bm{$\mathcal{Y}$}. The standard method for evaluating learning outcomes follows an \textit{empirical} risk/loss minimization formulation:\footnote{\scriptsize{We omit the regularization term for brevity.}}
\vspace{-3pt}
\begin{equation*}\small
	\tilde{f}\in\mathop{\arg\min}_{f\in\bm{\mathcal{F}}} \frac{1}{N}\sum\nolimits_n\varepsilon(f(x_n),z(x_n)),
	\vspace{-3pt}
\end{equation*}
where \bm{$\mathcal{F}$} denotes the hypothesis (solution) space, and $\varepsilon_{\!\!}:$ $\bm{\mathcal{X}}_{\!}\times\bm{\mathcal{Y}}_{\!}\mapsto\!\mathbb{R}$ is an error function that evaluates the estimate $f(x_n)$ against video segmentation related prior knowledge $z(x_n)\!\!\in$\bm{$\mathcal{Z}$}. To make $\tilde{f}$  a good approximation of $f^{*\!}$, current supervised video segmentation methods directly use the desired output $y_n$, \ie, $z(x_n)\!\!:=\!\!f^{*\!}(x_n)$, as the prior knowledge, {with$_{\!}$ the$_{\!}$ price$_{\!}$ of$_{\!}$ requiring$_{\!}$ vast$_{\!}$ amounts$_{\!}$ of$_{\!}$ well-labeled$_{\!}$ data.}

\noindent\red{$\bullet$~\textbf{Unsupervised (\red{Self-supervised}) Learning Methods.} When only data samples $\{x_n\}_{n\!\!}\!\subset$\bm{$\mathcal{X}$} are given, the problem of approximating $f^{*\!}$ is known as unsupervised learning. Unsupervised learning includes fully unsupervised learning methods in which
	the methods do not need any labels at all, as well as self-supervised learning methods in which networks are explicitly trained with automatically-generated pseudo labels without any human annotations~\cite{jing2020self}. Almost all existing unsupervised learning-based video segmentation models are self-supervised learning methods, where the prior knowledge \bm{$\mathcal{Z}$} refers to pseudo labels derived from intrinsic properties of video data (\eg, cross-frame consistency). We thus use ``unsupervised learning'' and ``self-supervised learning''  interchangeably. }

\noindent$\bullet$~\textbf{Weakly-Supervised Learning Methods.} In this case, \bm{$\mathcal{Z}$} is typically a more easily-annotated domain, {such as tags, bounding boxes, or scribbles,} and  $f^{*}$ is approximated using a finite number of samples from \bm{$\mathcal{X}$}$\times$\bm{$\mathcal{Z}$}.

\noindent\textbf{Remark}. So far, deep supervised learning-based methods are dominant in the field of video segmentation. However, exploring the task in an unsupervised or weakly supervised
setting is more appealing, not only because it alleviates the annotation burden of
\bm{$\mathcal{Y}$}, but because it inspires an in-depth$_{\!}$ understanding$_{\!}$ of$_{\!}$ the$_{\!}$ nature$_{\!}$ of$_{\!}$ the$_{\!}$ task$_{\!}$ by$_{\!}$ exploring$_{\!}$ \bm{$\mathcal{Z}$}.

\vspace{-3pt}
\subsection{History and Terminology}\label{sec:2.2}
Digital image segmentation has been studied for at least 50 years, starting with the Roberts
operator~\cite{roberts1965machine} for identifying object boundaries.
Since then, numerous algorithms~for image segmentation have been proposed, and many are extended to the video domain.  The field of video segmentation has evolved quickly and undergone great change.

Earlier attempts focus on \textbf{video over-segmentation}, \ie, partitioning a video into
space-time homogeneous, perceptually distinct-regions. Typical approaches include hierarchical video
segmentation~\cite{grundmann2010}, temporal superpixel~\cite{chang2013}, and super-voxels~\cite{xu2012evaluation}, based on the discontinuity and similarity of pixel intensities in a particular location, \ie, separating pixels according to abrupt changes in intensity or grouping pixels with similar intensity together.  These methods are instructive for early stage video preprocessing, but cannot solve the problem of object-level pattern modeling, as they do not provide any principled approach to flatten the hierarchical video decomposition into a binary segmentation~\cite{perazzi2015fully,DBLP:conf/iccv/PapazoglouF13}.

To extract foreground objects from video sequences, \textbf{background subtraction} techniques emerged {beginning in the late 70s~\cite{jain1979analysis}}, and became popular following the work of~\cite{wren1997pfinder}. They assume that the background is known a priori, and that the camera is stationary\!~\cite{criminisi2006bilayer,brutzer2011evaluation} or undergoes a predictable, parametric 2D\!~\cite{hayman2003statistical} or 3D motion with 3D parallax\!~\cite{DBLP:journals/pami/IraniA98}. These geometry-based methods fit well for specific application scenarios such as surveillance systems~\cite{brutzer2011evaluation,perazzi2015fully}, but they are sensitive to model selection (2D or 3D), and cannot handle non-rigid camera movements.

Another group of video segmentation solutions tackled the task of \textbf{motion segmentation}, \ie, finding objects in motion. Background subtraction can also be viewed as a specific case of motion segmentation. However, most motion segmentation models are built upon motion analysis~\cite{wang1993layered,sawhney1996compact}, factorization\!~\cite{costeira1995multi}, and/or statistical\!~\cite{cremers2005motion} techniques that comprehensively model the characteristics of moving scenes without prior knowledge of camera motion. Among the big family of motion segmentation algorithms, \textbf{trajectory segmentation} attained particular attention~\cite{Brox2010,DBLP:conf/iccv/OchsB11,DBLP:conf/cvpr/FragkiadakiZS12,DBLP:conf/iccv/KeuperAB15,DBLP:journals/pami/OchsMB14}. Trajectories are generated through tracking points over multiple frames and can represent long-term motion patterns, serving as an informative cue for segmentation.  Though impressive, motion-based methods heavily rely on the accuracy of optical flow estimation and can fail when different parts of an object exhibit heterogeneous motions.

To overcome these limitations, the task of extracting generic objects from unconstrained video sequences, \ie, AVOS, has drawn increasing research interest~\cite{DBLP:conf/bmvc/FaktorI14}. Several methods \cite{lee2011key,ma2012,zhang2013,xiao2016track} explored object hypotheses or proposals~\cite{endres2010category} as middle-level object representations. They generate a large number of object candidates in every frame and cast the task of segmenting video objects as an object region selection problem. The main drawbacks of the proposal-based algorithms are the high computational cost \cite{perazzi2016benchmark}  and complicated object inference schemes. Some others explored heuristic hypotheses such as visual attention~\cite{DBLP:conf/cvpr/WangSP15} and motion boundary~\cite{DBLP:conf/iccv/PapazoglouF13}, but  easily fail in scenarios where the heuristic assumptions do not hold.

As argued earlier, an alternative to the above unattended solutions is {to incorporate human-marked initialization}, \ie, SVOS. Older SVOS methods often rely on optical flow \cite{Ramakanth2014,shankar2015,Tsai2016,wang2017} and share a similar spirit with \textit{object tracking}~\cite{tsai2010,Wen_2015_CVPR}. In addition, some pioneering IVOS methods were proposed to address high-quality video segmentation under extensive human guidance, including rotoscoping~\cite{agarwala2004keyframe,li2016roto++}, scribble~\cite{Bai2009,Criminisi2010,zhong2012discontinuity,Fan2015,shankar2015}, contour~\cite{lu2016coherent}, and points~\cite{wang2017selective}. Significant engineering is typically needed to allow IVOS systems to operate at interactive speeds. In short, SVOS and IVOS pay for the improved flexibility and accuracy: they are infeasible at large scale due to their human-in-the-loop nature.

In the pre-deep learning era, {relatively few papers$_{\!}$~\cite{jain2013coarse,kae2014shape,tang2013discriminative,liu2014weakly,liu2015multiclass}  considered VSS due to the complexity of the task. The approaches typically relied on supervised classifiers such as SVMs and video over-segmentation} techniques.  

Overall, traditional approaches for video segmentation, though giving interesting results, are  constrained by hand-crafted features and heavy engineering. {But deep learning brought the performance of video segmentation to a new level, as we will review in \S\ref{sec:3}.}

\vspace{-4pt}
\subsection{Related Research Areas}\label{sec:2.3}
There are several research fields {closely related to video segmentation, which we now  briefly describe.}

\noindent$\bullet$~\textbf{Visual Tracking.} To infer the location of a target object over time, current tracking methods usually assume that the target is determined by a bounding box in the first frame~\cite{smeulders2013visual}. However, in more general tracking scenarios, and in particular the cases studied in early tracking methods, diverse object representations are explored~\cite{yilmaz2006object}, including centroids, skeletons, and contours. Some video segmentation techniques, such as background subtraction, are also merged into older trackers~\cite{bibby2008robust,ren2007tracking}. Hence, visual tracking and video segmentation encounter some common challenges (\eg, object/camera motion, appearance change, occlusion, \etc), fostering their mutual collaboration.

\noindent$\bullet$~\textbf{Image Semantic Segmentation.} The success of end-to-end image semantic segmentation~\cite{garcia2018survey,wang2021exploring,zhou2022rethinking} {has sparked the rapid development of VSS}. Rather than directly applying image semantic segmentation techniques frame by frame, recent VSS systems explore temporal continuity to increase both accuracy and efficiency. Nevertheless, image semantic segmentation techniques continue to serve as a foundation for advancing segmentation in video.

\begin{table*}
\centering
\caption{
	{\textbf{Summary of essential characteristics for reviewed AVOS methods} (\S\ref{sec:3.1.1}).
		\protect\\ \textbf{Instance}: instance- or object-level segmentation; \textbf{Flow}: if optical flow is used.
	}
}
\vspace{-6pt}
\label{table:AVOS_methods}
\begin{threeparttable}
	\resizebox{0.89\textwidth}{!}{
		\setlength\tabcolsep{6pt}
		\renewcommand\arraystretch{1.0}
		\begin{tabular}{|c|r||c|c|c|c|c|c|}
			\hline\thickhline
			\rowcolor{mygray}
			
			Year &Method~~~~  &Pub. &Core Architecture  
			& Instance & Flow 
			& Training Dataset \\
			\hline
			\hline
			\hline
			\multirow{5}{*}{\rotatebox{90}{2017}}
			& FSEG~\cite{jain2017fusionseg} & CVPR & Two-Stream FCN 
			& Object & \checkmark 
			& ImageNet VID~\cite{ILSVRC15} + DAVIS$_{16}$~\cite{perazzi2016benchmark} \\
			& SFL~\cite{cheng2017segflow} & ICCV & Two-Stream FCN 
			& Object& \checkmark 
			& DAVIS$_{16}$~\cite{perazzi2016benchmark} \\
			& LVO~\cite{DBLP:conf/iccv/TokmakovAS17} & ICCV & Two-Stream FCN 
			& Object & \checkmark 
			& DAVIS$_{16}$~\cite{perazzi2016benchmark} \\
			&LMP~\cite{DBLP:conf/cvpr/TokmakovAS17}  &ICCV &FCN 
			& Object & \checkmark 
			& FT3D~\cite{mayer2016large}\\
			&NRF~\cite{li2017primary} &ICCV &FCN 
			& Object& \checkmark  
			&Youtube-Objects~\cite{DBLP:conf/cvpr/PrestLCSF12} \\
			
			\hline
			\multirow{6}{*}{\rotatebox{90}{2018}}
			&IST~\cite{Li_2018_CVPR}  &CVPR &FCN 
			&Object & \checkmark 
			& DAVIS$_{16}$~\cite{perazzi2016benchmark} \\
			&FGRNE~\cite{li2018flow}&CVPR &FCN + RNN 
			& Object & 
			&SegTrackV2~\cite{li2013video} + DAVIS$_{16}$~\cite{perazzi2016benchmark} + FBMS~\cite{DBLP:journals/pami/OchsMB14} \\
			&MBN~\cite{Li_2018_ECCV1} &ECCV &FCN 
			& Object & \checkmark 
			& DAVIS$_{16}$~\cite{perazzi2016benchmark}\\
			&PDB~\cite{Song_2018_ECCV}&ECCV &RNN 
			& Object & 
			&DAVIS$_{16}$~\cite{perazzi2016benchmark} \\
			& MOT~\cite{siam2018video} &ICRA& Two-Stream FCN 
			& Object & \checkmark
			&DAVIS$_{16}$~\cite{perazzi2016benchmark}  \\
			
			\hline
			\multirow{9}{*}{\rotatebox{90}{2019}}
			&RVOS~\cite{ventura2019rvos} &CVPR &RNN 
			& Instance &  
			& DAVIS$_{17}$~\cite{pont20172017}/YouTube-VIS~\cite{yang2019video}\\
			&COSNet~\cite{Lu_2019_CVPR} & CVPR & Siamese FCN + Co-attention 
			& Object & 
			&  MSRA10K~\cite{cheng2015global} + DUT~\cite{DBLP:conf/cvpr/YangZLRY13} + DAVIS$_{16}$~\cite{perazzi2016benchmark}\\
			&UMOD~\cite{yang2019unsupervised} & CVPR & Adversarial Network 
			& Object & \checkmark 
			& SegTrackV2~\cite{li2013video} + DAVIS$_{16}$~\cite{perazzi2016benchmark} + FBMS~\cite{DBLP:journals/pami/OchsMB14} \\
			&AGS~\cite{wang2019learning2} & CVPR & FCN 
			& Object &  
			& SegTrackV2~\cite{li2013video} + DAVIS$_{16}$~\cite{perazzi2016benchmark} + DUT~\cite{DBLP:conf/cvpr/YangZLRY13} + PASCAL-S~\cite{li2014secrets} \\
			&AGNN~\cite{wang2019zero} &ICCV &FCN + GNN 
			& Object & 
			& MSRA10K~\cite{cheng2015global} + DUT~\cite{DBLP:conf/cvpr/YangZLRY13} + DAVIS$_{16}$~\cite{perazzi2016benchmark}\\
			&MGA~\cite{Li_2019_ICCV} &ICCV &Two-Stream FCN 
			& Object & \checkmark 
			& DUTS~\cite{wang2017learning} + DAVIS$_{16}$~\cite{perazzi2016benchmark} + FBMS~\cite{DBLP:journals/pami/OchsMB14}\\
			&AnDiff~\cite{yang2019anchor} & ICCV & Siamese FCN + Co-attention 
			& Object & 
			& DAVIS$_{16}$~\cite{perazzi2016benchmark}\\
			&LSMO~\cite{Tokmakov2019} &IJCV &Two-Stream FCN 
			& Object & \checkmark 
			& FT3D~\cite{mayer2016large} + DAVIS$_{16}$~\cite{perazzi2016benchmark}\\	
			\hline
			\multirow{7}{*}{\rotatebox{90}{2020}}
			&MATNet~\cite{zhou2020motion} &AAAI &Two-Stream FCN 
			& Object &\checkmark 
			& Youtube-VOS~\cite{xu2018youtube} + DAVIS$_{16}$~\cite{perazzi2016benchmark}\\
			&PyramidCSA~\cite{gu2020pyramid} &AAAI &Siamese FCN + Co-attention 
			& Object & 
			& DUTS~\cite{wang2017learning} + DAVIS$_{16}$~\cite{perazzi2016benchmark} + DAVSOD~\cite{fan2019shifting}\\
			& MuG~\cite{lu2020learning} & CVPR & FCN 
			& Object & 
			& OxUvA~\cite{valmadre2018long} \\
			&EGMN~\cite{lu2020video} & ECCV & FCN + Episodic Memory 
			& Object & 
			& MSRA10K~\cite{cheng2015global} + DUT~\cite{DBLP:conf/cvpr/YangZLRY13} + DAVIS$_{16}$~\cite{perazzi2016benchmark} \\
			&WCSNet~\cite{zhang2020unsupervised} & ECCV & Siamese FCN 
			& Object & 
			& SALICON~\cite{jiang2015salicon} + PASCAL VOC 2012~\cite{DBLP:journals/ijcv/EveringhamEGWWZ15}  + DUTS~\cite{wang2017learning} + DAVIS$_{16}$~\cite{perazzi2016benchmark} \\
			&DFNet~\cite{zhen2020learning} & ECCV & Siamese FCN 
			& Object & 
			& MSRA10K~\cite{cheng2015global} + DUT~\cite{DBLP:conf/cvpr/YangZLRY13} + DAVIS$_{16}$~\cite{perazzi2016benchmark} \\
			
			\hline
			\multirow{5}{*}{\rotatebox{90}{2021}}
			&F2Net~\cite{liu2021f2net} &AAAI &Siamese FCN 
			& Object &
			& MSRA10K~\cite{cheng2015global} + DAVIS$_{16}$~\cite{perazzi2016benchmark}\\
			&TODA~\cite{zhou2021target} & CVPR & Siamese FCN 
			& Instance & 
			& DAVIS$_{17}$~\cite{pont20172017}/YouTube-VIS~\cite{yang2019video} \\
			&RTNet~\cite{Ren_2021_CVPR} & CVPR & Two-Stream FCN 
			& Object & \checkmark 
			& DUTS~\cite{wang2017learning} + DAVIS$_{16}$~\cite{perazzi2016benchmark} \\
			&DyStab~\cite{yang2021dystab} & CVPR & Adversarial Network 
			& Object & \checkmark 
			& SegTrackV2~\cite{li2013video} + DAVIS$_{16}$~\cite{perazzi2016benchmark} + FBMS~\cite{DBLP:journals/pami/OchsMB14} \\
			
			& \red{MotionGrouping \cite{yang2021self}} & \red{ICCV} & \red{Transformer} & \red{Object} &  \red{\checkmark} & \red{DAVIS16~\cite{perazzi2016benchmark}/SegTrackV2~\cite{li2013video}/FBMS59~\cite{ochs2013segmentation}/MoCA~\cite{lamdouar2020betrayed}} \\
			
			\hline
		\end{tabular}
	}
\end{threeparttable}
\vspace{-10pt}
\end{table*}

\noindent$\bullet$~\textbf{Video Object Detection.} To generalize object detection in the video domain~\cite{jiao2021new}, video object detectors incorporate temporal cues over the box- or feature- level. There are many key technical steps and challenges, such as object proposal generation, temporal information aggregation, and cross-frame object association, that are shared between video object detection and (instance-level) video segmentation.

\vspace{-4pt}
\section{Deep Learning-based Video Segmentation}\label{sec:3}

\subsection{Deep Learning-based VOS Models}\label{sec:3.1}
VOS extracts generic foreground objects from video sequences with no concern for semantic category recognition. Based on how much {human intervention is involved in inference, VOS models can be divided into three classes (\S\ref{sec:2.1.2}): automatic (AVOS, \S\ref{sec:3.1.1}), semi-automatic (SVOS, \S\ref{sec:3.1.2}), and interactive (IVOS, \S\ref{sec:3.1.3}).} Moreover, although language-guided video object segmentation (LVOS) falls in the broader category of SVOS, LVOS methods are reviewed alone (\S\ref{sec:3.1.4}), due to the specific multi-modal task setup.

	\vspace{-4pt}
\subsubsection{Automatic Video Object Segmentation (AVOS)}\label{sec:3.1.1}
Instead of using heuristic priors and hand-crafted features to automatically execute VOS, modern AVOS methods learn {generic video object patterns in a data-driven fashion. We group landmark efforts based on their key techniques.}

\noindent$\bullet$~\textbf{Deep Learning Module based Methods.} In 2015, Fragkiadaki \etal~\cite{fragkiadaki2015learning} made an early effort that learns a multi-layer perceptron to rank proposal segments and infer foreground objects. In 2016, Tsai \etal~\cite{Tsai2016} proposed a joint optimization framework for AVOS and optical flow estimation with a na\"{i}ve use of deep features from a pre-trained classification network. Later methods~\cite{li2017primary,DBLP:conf/cvpr/TokmakovAS17} learn FCNs to predict initial, pixel-level foreground estimates from frame images~\cite{li2017primary,wang2017video} or optical flow fields~\cite{DBLP:conf/cvpr/TokmakovAS17}, while several post-processing steps are still needed. Basically, these primitive solutions largely rely on traditional AVOS techniques; the learning ability of neural networks is under-explored.

\noindent$\bullet$~\textbf{Pixel Instance Embedding based Methods.}  A group of AVOS models has been developed to make use of stronger deep learning descriptors~\cite{Li_2018_CVPR,Li_2018_ECCV1} -- instance embeddings -- learned from image instance segmentation data~\cite{fathi2017semantic}. \red{They first generate pixel-wise instance embeddings, and select representative embeddings which are clustered into foreground and background. Finally, the labels of the sampled embeddings are propagated to the other ones. The clustering and  propagation can be achieved without video specific supervision. Though using fewer annotations, these methods suffer from a fragmented and complicated pipeline.}

\noindent$\bullet$~\textbf{End-to-end Methods with Short-term Information Encoding.} End-to-end model designs became the mainstream in this field. For example, convolutional recurrent neural networks (RNNs) were used to learn spatial and temporal visual patterns jointly~\cite{Song_2018_ECCV,wang2019zero}. Another big family is built upon two-stream networks~\cite{jain2017fusionseg,cheng2017segflow,Li_2019_ICCV,li2018flow,Tokmakov2019,DBLP:conf/iccv/TokmakovAS17,zhou2020motion}, wherein two parallel streams
are built to extract features from raw image and optical flow, which are further fused for segmentation prediction. Two-stream methods make explicit use of appearance and motion cues, at the cost of optical flow computation and vast learnable parameters. {These end-to-end methods improve accuracy and show the advantages of applying neural
networks to this task. However, they only consider local content within very limited time span; they stack appearance and/or motion information from a few successive frames as input, ignoring relations among distant frames. Although RNNs are usually adopted, their internal hidden memory creates the inherent limits in modeling longer-term dependencies~\cite{sukhbaatar2015end}.}

\begin{table*}
	\centering
	\caption{
		{\textbf{Summary of essential characteristics for reviewed  SVOS methods} (\S\ref{sec:3.1.2}).
			\textbf{Flow}: if optical flow is used.
}
	}
	\vspace{-5pt}
	\label{table:SVOS_methods}
	\begin{threeparttable}
		\resizebox{0.99\textwidth}{!}{
			\setlength\tabcolsep{6pt}
			\renewcommand\arraystretch{1.0}
			\begin{tabular}{|c|r||c|c|c|c|c|}
				\hline\thickhline
				\rowcolor{mygray}
				
				Year &Method~~~~  &Pub. &Core Architecture 
& Flow 
& Technical Feature & Training Dataset \\
				\hline
				\hline

				\hline
				\multirow{7}{*}{\rotatebox{90}{2017}}
				&OSVOS~\cite{DBLP:conf/cvpr/CaellesMPLCG17} &CVPR & FCN 
& 
&  Online Fine-tuning&DAVIS$_{16}$~\cite{perazzi2016benchmark} \\
				& MaskTrack~\cite{DBLP:conf/cvpr/PerazziKBSS17} & CVPR & FCN 
& \checkmark 
& Propagation-based & ECSSD~\cite{yan2013hierarchical} + MSRA10K~\cite{cheng2015global} + PASCAL-S~\cite{li2014secrets} + DAVIS$_{16}$~\cite{perazzi2016benchmark}  \\
				& CTN~\cite{jang2017online} & CVPR & FCN 
& \checkmark 
& Propagation-based & PASCAL VOC 2012~\cite{DBLP:journals/ijcv/EveringhamEGWWZ15}  \\
				& VPN~\cite{DBLP:conf/cvpr/JampaniGG17} & CVPR & Bilateral Network 
& 
& Propagation-based  &DAVIS$_{16}$~\cite{perazzi2016benchmark} \\
				& PLM~\cite{DBLP:conf/iccv/YoonRKLSK17}& CVPR & Siamese FCN 
& 
& Matching-based & DAVIS$_{16}$~\cite{perazzi2016benchmark}\\
				&OnAVOS~\cite{voigtlaender2017online} &BMVC & FCN 
& 
& Online Fine-tuning &PASCAL VOC 2012~\cite{DBLP:journals/ijcv/EveringhamEGWWZ15} + COCO~\cite{lin2014microsoft}  + DAVIS~\cite{perazzi2016benchmark} \\
				& Lucid~\cite{DAVIS2017-2nd} & IJCV & Two-Stream FCN 
& \checkmark 
& Propagation-based & DAVIS$_{16}$~\cite{perazzi2016benchmark}\\
				\hline
				\multirow{12}{*}{\rotatebox{90}{2018}}
				& CINM~\cite{Bao_2018_CVPR} & CVPR & Spatio-temporal MRF  
& \checkmark 
& Propagation-based  & DAVIS$_{17}$~\cite{pont20172017} \\
				& FAVOS~\cite{cheng2018fast} & CVPR & FCN 
& 
& Propagation-based & DAVIS$_{16}$~\cite{perazzi2016benchmark}/DAVIS$_{17}$~\cite{pont20172017} \\
				& RGMP~\cite{wug2018fast} & CVPR & Siamese FCN 
& 
&Propagation-based & PASCAL VOC 2012~\cite{DBLP:journals/ijcv/EveringhamEGWWZ15} + ECSSD~\cite{yan2013hierarchical} + MSRA10K~\cite{cheng2015global} + DAVIS$_{17}$~\cite{pont20172017} \\
				& OSMN~\cite{yang2018efficient} & CVPR & FCN + Meta Learning  
&  
& Online Fine-tuning & ImageNet VID~\cite{ILSVRC15}  + DAVIS$_{16}$~\cite{perazzi2016benchmark} \\
				& MONet~\cite{xiao2018monet} & CVPR & FCN  
& \checkmark   
& Online Fine-tuning&PASCAL VOC 2012~\cite{DBLP:journals/ijcv/EveringhamEGWWZ15} + DAVIS$_{16}$~\cite{perazzi2016benchmark} \\
				& CRN~\cite{hu2018motion} & CVPR & FCN + Active Contour 
& \checkmark 
& Propagation-based & PASCAL VOC 2012~\cite{DBLP:journals/ijcv/EveringhamEGWWZ15} + DAVIS$_{16}$~\cite{perazzi2016benchmark} \\
				& RCAL~\cite{han2018reinforcement} & CVPR & FCN + RL 
& 
& Propagation-based & MSRA10K~\cite{cheng2015global} + PASCAL-S + SOD + ECSSD~\cite{yan2013hierarchical} + DAVIS$_{16}$~\cite{perazzi2016benchmark} \\
				&OSVOS-S~\cite{maninis2018video}  &PAMI & FCN 
&  
& Online Fine-tuning& DAVIS$_{16}$~\cite{perazzi2016benchmark}/DAVIS$_{17}$~\cite{pont20172017} \\
				& Videomatch~\cite{Hu_2018_ECCV} & ECCV & Siamese FCN 
&  
& Matching-based & DAVIS$_{16}$~\cite{perazzi2016benchmark}/DAVIS$_{17}$~\cite{pont20172017} \\
				& Dyenet~\cite{Li_2018_ECCV} & ECCV & Re-ID 
& 
& Propagation-based& DAVIS$_{17}$~\cite{pont20172017} \\
				& LSE~\cite{Ci_2018_ECCV} & ECCV & FCN 
& 
& Propagation-based& PASCAL VOC 2012~\cite{DBLP:journals/ijcv/EveringhamEGWWZ15} \\
				& Colorization~\cite{vondrick2018tracking} & ECCV & Siamese FCN 
& 
& Unsupervised Learning& Kinetics~\cite{kay2017kinetics} \\
				
				\hline
				\multirow{13}{*}{\rotatebox{90}{2019}}
				&MVOS~\cite{xiao2019online} & PAMI & Siamese FCN + Meta Learning 
& 
&Online Fine-tuning   &  PASCAL VOC 2012~\cite{DBLP:journals/ijcv/EveringhamEGWWZ15} + DAVIS$_{16}$~\cite{perazzi2016benchmark}/DAVIS$_{17}$~\cite{pont20172017}\\
				& FEELVOS~\cite{voigtlaender2019feelvos} & CVPR & FCN 
& 
& Matching-based& COCO~\cite{lin2014microsoft}  + DAVIS$_{17}$~\cite{pont20172017} + YouTube-VOS~\cite{xu2018youtube} \\
				& MHP-VOS~\cite{xu2019mhp} & CVPR & Graph Optimization 
& 
& Propagation-based & COCO~\cite{lin2014microsoft}  + DAVIS$_{16}$~\cite{perazzi2016benchmark}/DAVIS$_{17}$~\cite{pont20172017}\\
				& AGSS~\cite{lin2019agss} & CVPR & FCN 
& \checkmark
&  Propagation-based&  DAVIS$_{17}$~\cite{pont20172017}/YouTube-VOS~\cite{xu2018youtube} \\
				& AGAME~\cite{johnander2019generative} & CVPR & FCN 
& 
&   Propagation-based&  MSRA10K~\cite{cheng2015global} + PASCAL VOC 2012~\cite{DBLP:journals/ijcv/EveringhamEGWWZ15} + DAVIS$_{17}$~\cite{pont20172017}/YouTube-VOS~\cite{xu2018youtube} \\
				& SiamMask~\cite{wang2019fast} & CVPR & Siamese FCN 
& 
&  Box-Initialization&  DAVIS$_{16}$~\cite{perazzi2016benchmark}/DAVIS$_{17}$~\cite{pont20172017}/YouTube-VOS~\cite{xu2018youtube} \\
				&RVOS~\cite{ventura2019rvos} &CVPR & RNN 
& 
&  Propagation-based & DAVIS$_{17}$~\cite{pont20172017}/YouTube-VIS~\cite{yang2019video}\\
				&BubbleNet~\cite{griffin2019bubblenets} &CVPR & Siamese Network
&  
&  Bubble Sorting & DAVIS$_{17}$~\cite{pont20172017}\\
				&RANet~\cite{wang2019ranet} & ICCV & Siamese FCN 
& 
& Matching-based& MSRA10K~\cite{cheng2015global} + ECSSD~\cite{yan2013hierarchical}+ HKU-IS~\cite{li2015visual} + DAVIS$_{16}$~\cite{perazzi2016benchmark}/DAVIS$_{17}$~\cite{pont20172017} \\
				& DMM-Net~\cite{Zeng_2019_ICCV} & ICCV & Mask R-CNN 
& 
&Differentiable Matching &  DAVIS$_{17}$~\cite{pont20172017}/YouTube-VOS~\cite{xu2018youtube} \\
				& DTN~\cite{zhang2019fast} & ICCV & FCN 
& \checkmark 
&  Propagation-based& COCO~\cite{lin2014microsoft}  + PASCAL VOC 2012~\cite{DBLP:journals/ijcv/EveringhamEGWWZ15} + DAVIS$_{16}$/DAVIS$_{17}$~\cite{pont20172017}\\
				& STM~\cite{Oh_2019_ICCV} & ICCV &  Memory Network 
& 
& Matching-based&  PASCAL VOC 2012~\cite{DBLP:journals/ijcv/EveringhamEGWWZ15} + COCO~\cite{lin2014microsoft}  + ECSSD~\cite{yan2013hierarchical} + DAVIS$_{17}$~\cite{pont20172017}/YouTube-VOS~\cite{xu2018youtube}\\
				& TimeCycle~\cite{wang2019learning3} & ECCV & Siamese FCN 
& 
& Unsupervised Learning& VLOG~\cite{fouhey2018lifestyle} \\
				&{UVC} \cite{li2019joint} &{NeurIPS} & {Siamese FCN} 
& 
& {Unsupervised Learning} & {COCO~\cite{lin2014microsoft} + Kinetics~\cite{kay2017kinetics}}\\
				
				\hline
				\multirow{16}{*}{\rotatebox{90}{2020}}
				&e-OSVOS~\cite{meinhardt2020make} & NeurIPS &  Mask R-CNN + Meta Learning
&   
& Online Fine-tuning&  DAVIS$_{17}$~\cite{pont20172017} + YouTube-VOS~\cite{xu2018youtube}\\
				& AFB-URR~\cite{liang2020video} & NeurIPS & Memory Network 
& 
& Matching-based& PASCAL VOC 2012~\cite{DBLP:journals/ijcv/EveringhamEGWWZ15} + COCO~\cite{lin2014microsoft} + ECSSD~\cite{yan2013hierarchical} + DAVIS$_{17}$~\cite{pont20172017}/YouTube-VOS~\cite{xu2018youtube}\\
				& Fasttan~\cite{huang2020fast} & CVPR & Faster R-CNN 
&  
& Propagation-based &  COCO~\cite{lin2014microsoft}  + DAVIS$_{17}$~\cite{pont20172017}\\
				& Fasttmu~\cite{sun2020fast} & CVPR  & FCN + RL 
& 
& Box-Initialization & PASCAL VOC 2012~\cite{DBLP:journals/ijcv/EveringhamEGWWZ15} + DAVIS$_{17}$~\cite{pont20172017}\\
				& SAT~\cite{chen2020state} & CVPR & FCN + RL 
& 
& Propagation-based& COCO~\cite{lin2014microsoft}  + DAVIS$_{17}$~\cite{pont20172017} + YouTube-VOS~\cite{xu2018youtube}\\
				& FRTM-VOS~\cite{robinson2020learning} & CVPR & FCN  
& 
& Matching-based&  DAVIS$_{17}$~\cite{pont20172017}/YouTube-VOS~\cite{xu2018youtube}\\
				& TVOS~\cite{zhang2020transductive} & CVPR & FCN  
& 
&  Matching-based& DAVIS$_{17}$~\cite{pont20172017}/YouTube-VOS~\cite{xu2018youtube}\\
				& MuG~\cite{lu2020learning} & CVPR & Siamese FCN  
& 
&  Unsupervised Learning& OxUvA~\cite{valmadre2018long}\\
				& MAST~\cite{lai2020mast} & CVPR & Memory Network  
& 
&  Unsupervised Learning& OxUvA~\cite{valmadre2018long} + YouTube-VOS~\cite{xu2018youtube}\\
				& GCNet~\cite{li2020fast} & ECCV & Memory Network 
& 
& Matching-based& MSRA10K~\cite{cheng2015global} + ECSSD~\cite{yan2013hierarchical} + HKU-IS~\cite{li2015visual} + DAVIS$_{17}$~\cite{pont20172017}/YouTube-VOS~\cite{xu2018youtube}\\
				& KMN~\cite{seong2020kernelized} & ECCV & Memory Network 
& 
&Matching-based & PASCAL VOC 2012~\cite{DBLP:journals/ijcv/EveringhamEGWWZ15} + COCO~\cite{lin2014microsoft}  + ECSSD~\cite{yan2013hierarchical} + DAVIS$_{17}$~\cite{pont20172017}/YouTube-VOS~\cite{xu2018youtube}\\
				& CFBI~\cite{yang2020collaborative} & ECCV & FCN 
& 
& Matching-based& COCO~\cite{lin2014microsoft} + DAVIS$_{17}$~\cite{pont20172017}/YouTube-VOS~\cite{xu2018youtube}\\
				& LWL~\cite{bhat2020learning} & ECCV & Siamese FCN + Meta Learning 
& 
& Matching-based&  DAVIS$_{17}$~\cite{pont20172017} + YouTube-VOS~\cite{xu2018youtube}\\
				& MSN~\cite{wu2020memory} & ECCV & Memory Network 
& 
& Matching-based&  DAVIS$_{17}$~\cite{pont20172017}/YouTube-VOS~\cite{xu2018youtube}\\
				& EGMN~\cite{lu2020video} & ECCV & Memory Network 
& 
& Matching-based&  MSRA10K~\cite{cheng2015global} + COCO~\cite{lin2014microsoft} + DAVIS$_{17}$~\cite{pont20172017} + YouTube-VOS~\cite{xu2018youtube}\\
				& STM-Cycle~\cite{li2020delving} & NeurIPS & Memory Network 
& 
& Matching-based&  DAVIS$_{17}$~\cite{pont20172017} + YouTube-VOS~\cite{xu2018youtube}\\
				\hline
				\multirow{8}{*}{\rotatebox{90}{2021}}
				& QMA~\cite{lin2021query} & AAAI & Memory Network 
& 
& Box-Initialization & DUT~\cite{DBLP:conf/cvpr/YangZLRY13} + HKU-IS~\cite{li2015visual} + MSRA10K~\cite{cheng2015global} + YouTube-VOS~\cite{xu2018youtube}\\
				& SwiftNet~\cite{Wang_2021_SwiftNetCVPR} & CVPR & Memory Network 
& 
& Matching-based&  COCO~\cite{lin2014microsoft} + DAVIS$_{17}$~\cite{pont20172017}/YouTube-VOS~\cite{xu2018youtube}\\
				& G-FRTM~\cite{Park_2021_CVPR} & CVPR & FCN + RL 
& 
& Matching-based&  DAVIS$_{17}$~\cite{pont20172017} + YouTube-VOS~\cite{xu2018youtube}\\
				& SST~\cite{duke2021sstvos} & CVPR & Transformer 
& 
& Matching-based&  DAVIS$_{17}$~\cite{pont20172017} + YouTube-VOS~\cite{xu2018youtube}\\
				& GIEL~\cite{Ge_2021_CVPR} & CVPR & Siamese FCN 
& 
& Matching-based&  DAVIS$_{17}$~\cite{pont20172017} + YouTube-VOS~\cite{xu2018youtube}\\
				& LCM~\cite{Hu_2021_CVPR} & CVPR & Memory Network  
& 
&Matching-based &  PASCAL VOC 2012~\cite{DBLP:journals/ijcv/EveringhamEGWWZ15} + COCO~\cite{lin2014microsoft}  + ECSSD~\cite{yan2013hierarchical} + DAVIS$_{17}$~\cite{pont20172017}/YouTube-VOS~\cite{xu2018youtube}\\
				& RMNet~\cite{Xie_2021_CVPR} & CVPR & Memory Network  
& \checkmark 
& Matching-based&  PASCAL VOC 2012~\cite{DBLP:journals/ijcv/EveringhamEGWWZ15} + COCO~\cite{lin2014microsoft}  + ECSSD~\cite{yan2013hierarchical} + DAVIS$_{17}$~\cite{pont20172017}/YouTube-VOS~\cite{xu2018youtube}\\
				& {CRW}~\cite{jabri2020space} & {NeurIPS} & {FCN}
& 
& {Unsupervised Learning} & {Kinetics} \cite{kay2017kinetics} \\

				\hline
			\end{tabular}
		}
	\end{threeparttable}
	\vspace{-10pt}
\end{table*}

\noindent$\bullet$~\textbf{End-to-end Methods with Long-term Context Encoding.} Current leading AVOS models use global context over long time spans. \red{In a seminal work~\cite{Lu_2019_CVPR}, Lu \etal proposed~a Siamese architecture-based  model that extracts features for arbitrary frame pairs and captures cross-frame context by calculating pixel-wise feature correlations. During inference, for each test frame, context from several other frames (within the same
video) is aggregated to locate objects. A contemporary work~\cite{yang2019anchor} exploited a similar idea but only used the first frame as reference. Several papers~\cite{zhang2020unsupervised,Ren_2021_CVPR} extended \cite{Lu_2019_CVPR} by making better use of information from multiple frames \cite{wang2019zero,lu2020zero,lu2021segmenting}, encoding spatial context \cite{liu2021f2net}, and incorporating temporal consistency to improve representation power and computation efficiency~\cite{gu2020pyramid,zhen2020learning}.}

\noindent$\bullet$~$_{\!}$\textbf{Un-/Weakly-supervised$_{\!}$ based$_{\!}$ Methods.} Only a handful of methods learn to perform AVOS from unlabeled or weakly labeled data. In~\cite{wang2019learning2}, static image salient object segmentation and dynamic eye fixation data, which are more easily acquired compared with video segmentation data, are used to learn video generic object patterns.  In~\cite{lu2020learning}, visual patterns are learned through exploring several intrinsic properties of video data at multiple granularities, \ie, intra-frame saliency, short-term visual coherence, long-range semantic correspondence, and video-level discriminativeness. In~\cite{yang2019unsupervised}, an adversarial contextual model is developed to segment  moving objects without any manual
annotation, achieved by minimizing the mutual information between the motions of an object and its context. This method is further enhanced in~\cite{yang2021dystab} by adopting a bootstrapping strategy and enforcing temporal consistency. \red{In~\cite{yang2021self}, motion is exclusively exploited to discover moving objects, and a Transformer-based model is designed and trained by self-supervised flow reconstruction using unlabeled video data.}

\noindent$\bullet$~\textbf{Instance-level AVOS Methods.} Instance-level AVOS, also referred as \textit{multi-object unsupervised video segmentation}, was introduced with the launch of the DAVIS$_{19}$ challenge~\cite{Caelles_arXiv_2019}. This task setting is more challenging as it requires not only separating the foreground objects from the background, but also discriminating different object instances. To tackle this task, current solutions typically work in a {top-down fashion}, \ie, generating object candidates for each frames, and associating instances over different frames. In an early attempt~\cite{ventura2019rvos}, Ventura~\etal delivered a recurrent network-based model that consists of a spatial LSTM for {per-frame instance discovery} and a temporal LSTM for cross-frame instance association. {This method features an elegant model design}, while its representation ability is too weak to enumerate all the object instances and to capture complex interactions between instances over the temporal domain. \red{Thus later methods~\cite{luiten2020unovost,wang2019learning,lu2020zero} strengthen the two-step pipeline through: \textbf{i)} employing image instance segmentation models (\eg, Mask R-CNN~\cite{he2017mask}) to detect object candidates, and \textbf{ii)} leveraging  tracking/re-identification techniques and manually designed rules for instance association. Foreground/background AVOS techniques~\cite{wang2019zero,Lu_2019_CVPR} are also used to filter out nonsalient candidates~\cite{wang2019learning,lu2020zero}. More recent methods, \eg, \cite{zhou2021target}, generate object candidates first and obtain corresponding tracklets via advanced SVOS techniques.} Overall, current instance-level AVOS models follow the classic tracking-by-detection paradigm, involving several ad-hoc designs. There is still considerable room for further improvement in accuracy and efficiency.

\subsubsection{Semi-automatic Video Object Segmentation (SVOS)}\label{sec:3.1.2}
Deep learning-based SVOS methods mainly focus on the first-frame \textit{mask} propagation setting.  They are categorized by their utilization of the test-time provided object masks.

\noindent$\bullet$~\textbf{Online Fine-tuning based Methods.} Following the one-

\noindent shot principle, this family of methods~\cite{DBLP:conf/cvpr/CaellesMPLCG17,maninis2018video,xiao2018monet,xiao2019online} trains a segmentation model separately on each given object mask in an online fashion. Fine-tuning methods essentially exploit the transfer learning capabilities of neural networks and often follow a two-step training procedure: i) \textit{offline pre-training}: learn general segmentation features from images and video sequences, and ii)
\textit{online fine-tuning}: learn target-specific representations from test-time supervision. The idea of fine-tuning was first introduced in~\cite{DBLP:conf/cvpr/CaellesMPLCG17}, where only the initial image-mask pair is used for {training an online, one-shot, but merely appearance-based FCN model.} Then, in~\cite{voigtlaender2017online}, more pixel samples in the unlabeled frames are mined as online training samples to better adapt to further changes over time.  As \cite{DBLP:conf/cvpr/CaellesMPLCG17,voigtlaender2017online} have no notion of individual objects,  \cite{maninis2018video} further incorporates instance segmentation models (\eg, Mask R-CNN~\cite{he2017mask}) during inference.  
While elegant through their simplicity, {fine-tuning methods have several weaknesses}: i) pre-training is fixed and not optimized for
subsequent fine-tuning, ii) hyperparameters of online fine-tuning are often excessively
hand-crafted and fail to generalize between test cases, iii) the common existing fine-tuning setups suffer from high test runtimes (up to 1,000 training iterations per segmented object online~\cite{DBLP:conf/cvpr/CaellesMPLCG17}). The root cause is that these approaches choose to encode all the target-related cues (\ie, appearance, mask) into network parameters implicitly. Towards efficient and automated fine-tuning, some recent methods~\cite{yang2018efficient,xiao2019online,meinhardt2020make} turn to meta learning techniques, \ie, optimize the fine-tuning policies~\cite{xiao2019online,meinhardt2020make} (\eg, generic model initialization, learning rates, \etc) or even directly modify network weights~\cite{yang2018efficient}.

\noindent$\bullet$~\textbf{Propagation-based Methods.} {Two recent lines of research\\
-- built upon mask propagation and template matching~techniques respectively -- try to refrain from the online~optimization to deliver compact, end-to-end SVOS solutions.} In particular, propagation-based methods use the previous frame mask to infer the current mask$_{\!}$~\cite{DBLP:conf/cvpr/PerazziKBSS17,jang2017online,DBLP:conf/cvpr/JampaniGG17}. For example, Jampani \etal~\cite{DBLP:conf/cvpr/JampaniGG17} propose a bilateral network for long-range
video-adaptive mask propagation. \red{Perazzi \etal\\ \cite{DBLP:conf/cvpr/PerazziKBSS17}} \red{approach SVOS by employing a modified FCN, where the previous frame mask is considered as an extra input channel. Follow-up work adopts optical flow guided mask alignment~\cite{xiao2018monet},  heavy first-frame data augmentation$_{\!}$~\cite{DAVIS2017-2nd}, and  multi-step segmentation refinement$_{\!}$~\cite{hu2018motion}. Others apply\\ re-identification to retrieve missing objects after prolonged occlusions$_{\!}$~\cite{Li_2018_ECCV}, design a reinforcement learning agent that tackles$_{\!}$ SVOS as$_{\!}$ a$_{\!}$ conditional$_{\!}$ decision-making$_{\!}$ process$_{\!}$~\cite{han2018reinforcement}, or propagate masks in a spatiotemporal MRF model~to~improve temporal coherency~\cite{Bao_2018_CVPR}. Some researchers leverage location-aware embeddings to sharpen the feature$_{\!}$~\cite{Ci_2018_ECCV}, or directly learn sequence-to-sequence mask propagation~\cite{xu2018youtube}.} Advanced tracking techniques are also exploited in~\cite{cheng2018fast,xu2019mhp,chen2020state,huang2020fast}. Propagation-based methods are found~to easily suffer from error accumulation due to occlusions~and drifts during mask propagation. Conditioning propagation on the initial frame-mask pair$_{\!}$~\cite{wug2018fast,lin2019agss,zhang2019fast} seems a feasible$_{\!}$ solution$_{\!}$ to$_{\!}$ this.$_{\!}$ Although$_{\!}$ target-specific$_{\!}$ mask$_{\!}$ is$_{\!}$ exp-\\ licitly encoded into the segmentation network, making up for$_{\!}$ the$_{\!}$ deficiencies$_{\!}$ of$_{\!}$ fine-tuning$_{\!}$ methods$_{\!}$ to$_{\!}$ a$_{\!}$ certain$_{\!}$ extent,\\ propagation-based$_{\!}$ methods still embed object appearance into hidden network weights. Clearly, {such implicit target-appearance modeling strategy hurts flexibility and adaptivity (\red{while$_{\!}$~\cite{johnander2019generative} is an exception -- a generative model of target\\ and background is explicitly built to aid mask propagation}).}

\begin{figure*}
	\begin{minipage}{\textwidth}
		{
			\begin{minipage}[t]{0.3\textwidth}
				{\normalsize tionally considers first-frame annotations, but easily fails in challenging scenes {without human feedback}. Moreover, the first-frame annotations are typically detailed masks, which are tedious to acquire: 79 seconds per instance for coarse polygon annotations of COCO~\cite{lin2014microsoft}, and much more for higher quality. Thus performing~ VOS~ in~ the~ interactive}
			\end{minipage}
		}
		\begin{minipage}[t]{0.7\textwidth}
			\vspace{-3ex}
			\makeatletter\def\@captype{table}\makeatother
			\caption{
				{\textbf{Summary of essential characteristics for reviewed  IVOS methods} (\S\ref{sec:3.1.3}).}
			}
	      \vspace{-5pt}
			\begin{threeparttable}[t]
				\resizebox{\textwidth}{!}{
					\setlength\tabcolsep{3pt}
					\renewcommand\arraystretch{1.0}
					\begin{tabular}{|c|r||c|c|c|c|c|c|}
						\hline\thickhline
						\rowcolor{mygray}
						Year &Method~~~~  &Pub. &Core Architecture  
						& Technical Feature & Training Dataset \\
						\hline
						\hline
						
						\hline
						2017
						& IIW~\cite{benard2017interactive} & - & FCN 
						& Interaction-Propagation  &PASCAL VOC 2012~\cite{DBLP:journals/ijcv/EveringhamEGWWZ15}  \\
						
						\hline
						2018
						& BFVOS~\cite{chen2018blazingly} & CVPR & FCN 
						& Pixel-wise Retrieval  &DAVIS$_{16}$~\cite{perazzi2016benchmark}  \\

						\hline
						2019
						&IVS~\cite{oh2019fast} &CVPR & FCN 
						& Interaction-Propagation & DAVIS$_{17}$~\cite{pont20172017}+YouTube-VOS~\cite{xu2018youtube}\\
						\hline
						\multirow{3}{*}{{2020}}
						& MANet~\cite{miao2020memory} & CVPR & Siamese FCN 
						& Interaction-Propagation & DAVIS$_{17}$~\cite{pont20172017} \\
						&ATNet~\cite{heo2020interactive} &ECCV & FCN 
						&  Interaction-Propagation& SBD + DAVIS$_{17}$~\cite{pont20172017}+YouTube-VOS~\cite{xu2018youtube}\\
						& ScribbleBox~\cite{chen2020scribblebox} & ECCV & GCN
						& Interaction-Propagation & COCO~\cite{lin2014microsoft} + ImageNet VID~\cite{ILSVRC15}  + YouTube-VOS~\cite{xu2018youtube} \\
						
						\hline
						\multirow{3}{*}{{2021}}
						& IVOS-W~\cite{Yin_2021_CVPR} & CVPR & FCN + RL 
						& Keyframe Selection & DAVIS$_{17}$~\cite{pont20172017} \\
						& GIS~\cite{Heo_2021_CVPR} & CVPR &  FCN 
						& Interaction-Propagation & DAVIS$_{17}$~\cite{pont20172017}+YouTube-VOS~\cite{xu2018youtube} \\
						& MiVOS~\cite{Cheng_2021_CVPRIVS} & CVPR & Memory Network 
						& Interaction-Propagation & BL30K~\cite{Cheng_2021_CVPRIVS}+DAVIS$_{17}$~\cite{pont20172017} + YouTube-VOS~\cite{xu2018youtube}\\
						
						\hline
				\end{tabular}}
				\vfill
				\vspace{-8pt}
			\end{threeparttable}
		\end{minipage}
	\end{minipage}
	\vspace*{-20pt}
\end{figure*}

\noindent$\bullet$$_{\!}$~\textbf{Matching-based$_{\!}$ Methods.$_{\!}$} \red{This type of methods, might the most promising SVOS solution so far, constructs an embedding space to memorize the initial object embeddings, and classifies each pixel's label according to their similarities to the target object in the embedding space.} Thus the initial object appearance is explicitly modeled, and test-time fine-tuning is not needed. The earliest effort in this direction can be tracked back to$_{\!}$~\cite{DBLP:conf/iccv/YoonRKLSK17}. Inspired by the advance in visual tracking~\cite{wang2015visual}, Yoon \etal~\cite{DBLP:conf/iccv/YoonRKLSK17} proposed a Siamese network to perform pixel-level matching between the first frame and\\ upcoming frames. Later, \cite{cheng2018fast} proposed to learn an embedding space from the first-frame supervision and pose VOS as a task of pixel retrieval: pixels are simply their respective nearest neighbors in the learned embedding space. The idea of~\cite{DBLP:conf/iccv/YoonRKLSK17} is also explored in~\cite{Hu_2018_ECCV}, while it computes two ma-\\ tching maps for each upcoming frame, with respect to the foreground and background annotated in the first frame.~In \cite{voigtlaender2019feelvos}, pixel-level similarities computed from the first frame and from the previous frame are used as a guide to segment succeeding frames. {Later, many matching-based solutions were proposed~\cite{wang2019ranet,Duarte_2019_ICCV}, perhaps most
notably Oh~\etal, who propose a space-time memory (STM) model to explicitly store previously computed segmentation information in an external memory~\cite{Oh_2019_ICCV}.} The memory facilitates learning the evolution of objects over time and allows for comprehensive use of past segmentation cues even over long period of time. Almost all current top-leading SVOS solutions~\cite{zhang2020transductive,yang2020collaborative} are built upon STM; they improve the target adaption ability~\cite{robinson2020learning,lu2020video,bhat2020learning}, incorporate local temporal continuity~\cite{seong2020kernelized,Xie_2021_CVPR,Hu_2021_CVPR}, explore instance-aware cues~\cite{Ge_2021_CVPR}, and develop more efficient memory designs~\cite{wu2020memory,li2020fast,liang2020video,Wang_2021_SwiftNetCVPR}. Recently, \cite{duke2021sstvos} introduced a Transformer~\cite{vaswani2017attention} based model, which performs matching-like computation through attending over a
history of multiple frames. In general, matching-based solutions enjoy the advantage of flexible and differentiable model design as well as long-term correspondence modeling.  On the
other hand, feature matching relies on a powerful and generic feature embedding, which may limit its performance in challenging scenarios.

It is also worth mentioning that, as an effective technique for target-specific model learning, online learning is applied by many propagation~\cite{DBLP:conf/cvpr/PerazziKBSS17,hu2018motion,Bao_2018_CVPR,xu2018youtube,xu2019mhp} and matching~\cite{DBLP:conf/iccv/YoonRKLSK17,wang2019ranet,Li_2018_ECCV} methods to boost performance.

\noindent$\bullet$~\textbf{Box-initialization based Methods.} As pixel-wise annotations are time-consuming or even impractical to acquire in realistic scenes, {some work has considered the situation where the first-frame annotation is provided in the form of a bounding box.  Specifically,} in~\cite{wang2019fast}, Siamese trackers~are augmented with a mask prediction branch. In~\cite{sun2020fast}, reinforcement learning is introduced to make decisions for target updating and matching. Later, in~\cite{lin2021query}, an outside memory is utilized to build a stronger Siamese track-segmenter.

\noindent$\bullet$~\textbf{Un-/Weakly-supervised based Methods.} To alleviate the demand for large-scale, pixel-wise annotated training samples, several un-/weakly-supervised learning-based SVOS solutions were recently developed. {They are typically built as$_{\!}$ a$_{\!}$ \textit{reconstruction}$_{\!}$ scheme$_{\!}$ (\ie,$_{\!}$ each$_{\!}$ pixel$_{\!}$ from$_{\!}$ a$_{\!}$ `query' frame is reconstructed by finding and assembling related pixels from adjacent frame(s)) \cite{vondrick2018tracking,lai2020mast,li2022locality}, and/or adopt a \textit{cycle-consistent tracking} paradigm (\ie, pixels/patches are encouraged to fall into the same location after one cycle of forward and backward tracking) \cite{wang2019learning3,li2019joint,lu2020learning,jabri2020space}.}

\noindent$\bullet$~\textbf{Other Specific Methods.} Other papers make specific contributions that deserve a separate look. In~\cite{Zeng_2019_ICCV}, Zeng~\etal extract mask proposals per frame and formulate the matching between object templates and proposals in a \textit{differentiable} manner. Instead of using only the first frame annotation, \cite{griffin2019bubblenets} learns to select \textit{the best frame} from the whole video for user interaction, so as to boost mask propagation. In~\cite{li2020delving}, Li~\etal introduce a forward-backward data flow based cycle consistency mechanism to improve both traditional SVOS training and offline inference protocols, through mitigating the error propagation problem. To accelerate processing speed, a dynamic network~\cite{Park_2021_CVPR} is proposed to selectively allocate computation source for each frame according to the similarity to the previous frame.

\subsubsection{Interactive Video Object Segmentation (IVOS)}\label{sec:3.1.3}
AVOS, without any human involvement, loses flexibility in segmenting arbitrary objects of user interest. SVOS addi-  { setting has gained increasing attention.  Unlike classic models \cite{li2016roto++,Bai2009,Fan2015} requiring extensive and professional user intervention, recent deep learning-based IVOS solutions usually work with multiple rounds of scribble supervision, to minimize the user's effort.} In this scenario~\cite{Caelles_arXiv_2018}, the user draws scribbles on a selected frame
and an algorithm computes the segmentation maps for all video frames in a batch process. For refinement, user intervention and segmentation are repeated. This \textit{round-based interaction}~\cite{oh2019fast} is useful~for consumer-level applications and rapid prototyping for professional usage, where efficiency is the main concern. {One can control the segmentation quality at the expense of time, as more rounds of interaction will provide better results.}

\noindent$\bullet$~\textbf{Interaction-propagation based Methods.} The majority of current studies~\cite{Heo_2021_CVPR,Cheng_2021_CVPRIVS} follow an \textit{interaction-propagation} scheme. \red{In the preliminary attempt~\cite{benard2017interactive}, IVOS is achieved by a simple combination of two separate modules: an interactive image segmentation model~\cite{xu2016deep} for producing segmentation based on user annotations; and a SVOS model~\cite{DBLP:conf/cvpr/CaellesMPLCG17} for propagating masks from the user-annotated frames to the others. Later, \cite{oh2019fast} devised a more compact solution, with also two modules for interaction and propagation, respectively.  However, the two modules are internally connected through intermediate feature exchanging, and also externally connected, \ie, each of them is conditioned on the other's output. In~\cite{heo2020interactive}, a similar model design is also adopted, however, the propagation part is specifically designed to address both local mask tracking (over adjacent frames) and global propagation (among distant frames), respectively.} {However, these
techniques \cite{benard2017interactive,heo2020interactive} have to start a new feed-forward computation in each interaction round, making them inefficient as the number of rounds grows.} A more efficient solution was developed in~\cite{miao2020memory}. The  critical idea is to build a common encoder for discriminative pixel embedding learning, upon which two small network branches are added for interactive segmentation and mask propagation, respectively.  Thus the model extracts pixel embeddings for all frames only once (in the first round). In the following rounds, the feed-forward computation is only made within the two shallow branches.

\noindent$\bullet$~\textbf{Other Methods.}  Chen \etal~\cite{chen2018blazingly} propose a pixel embedding learning-based model, applicable to both SVOS and IVOS. With a similar idea of~\cite{cheng2018fast}, IVOS is formulated as a pixel-wise retrieval problem, \ie, transferring labels to each pixel according to its nearest reference pixel. This model supports different kinds of user input, such as masks, clicks and scribbles, and can provide immediate feedback after user interaction. In~\cite{chen2020scribblebox}, an interactive annotation tool is proposed for VOS. The annotation has two phases: annotating objects with tracked boxes, and labeling masks inside these tracks.  Box tracks are annotated efficiently by approximating the trajectory using a parametric curve with a small number of control points which the annotator can interactively correct. Segmentation masks are corrected via scribbles which are propagated through time. In~\cite{Yin_2021_CVPR}, a reinforcement learning framework is exploited to automatically determine the most valuable frame for interaction.

\begin{table}
	\centering
\vspace{-5pt}
	\caption{
		\!\!{\textbf{Summary of characteristics for reviewed  LVOS methods} (\S\ref{sec:3.1.4}).}\!\!\!\!
	}
	\vspace{-7pt}
	\label{table:LVOS_methods}
	\begin{threeparttable}
		\resizebox{0.49\textwidth}{!}{
			\setlength\tabcolsep{3pt}
			\renewcommand\arraystretch{1.0}
			\begin{tabular}{|c|r||c|c|c|c|c|c|}
				\hline\thickhline
				\rowcolor{mygray}
				
				Year &Method~~~~  &Pub. &\tabincell{c}{Visual + Language\\Encoder}  &Technical Feature & Training Dataset \\
				\hline
				\hline
				
				\hline
				\multirow{2}{*}{{2018}}
				& A2DS~\cite{gavrilyuk2018actor} & CVPR & I3D + CNN &   Dynamic Conv. &A2D Sentences~\cite{gavrilyuk2018actor} \\
				& LangVOS~\cite{khoreva2018video} & ACCV & CNN + CNN &   Cross-modal Att. &DAVIS$_{17}$-RVOS~\cite{gavrilyuk2018actor} \\
				
				\hline
				\multirow{1}{*}{{2019}}
				& AAN~\cite{wang2019asymmetric} & ICCV & I3D + CNN &Cross-modal Att. & A2D Sentences~\cite{gavrilyuk2018actor} \\
				
				\hline
				\multirow{4}{*}{{2020}}
				& CDNet~\cite{wang2020context} & AAAI & I3D +  GRU & Dynamic Conv. & A2D Sentences~\cite{gavrilyuk2018actor} \\
				& PolarRPE~\cite{ningpolar} & IJCAI & I3D + LSTM  & Dynamic Conv. & A2D Sentences~\cite{gavrilyuk2018actor} \\
				& VT-Capsule~\cite{mcintosh2020visual} & CVPR & I3D + CNN  & Capsule Routing & A2D Sentences~\cite{gavrilyuk2018actor} \\
				& URVOS~\cite{seo2020urvos} & ECCV & CNN + MLP  & Cross-modal Att. & Refer-YouTube-VOS~\cite{seo2020urvos} \\
				
				\hline
				\multirow{2}{*}{{2021}}
				& CST~\cite{hui2021collaborative} & CVPR & I3D + GRU  & Cross-modal Att. & A2D Sentences~\cite{gavrilyuk2018actor} \\
				& CMSANet~\cite{ye2021referring} & PAMI & CNN + Word embed.  & Cross-modal Att. & A2D Sentences~\cite{gavrilyuk2018actor} \\

				\hline
			\end{tabular}
		}
	\end{threeparttable}
	\vspace{-12pt}
\end{table}

\begin{table*}
	\vspace{2pt}
	\centering
	\caption{%
		{\textbf{Summary of essential characteristics for reviewed VSS methods} (\S\ref{sec:3.2}). \textbf{Flow} indicates whether optical flow is used.}
	}
	\vspace{-5pt}
	\label{table:VSS_methods}
	\begin{threeparttable}
		\resizebox{0.94\textwidth}{!}{
			\setlength\tabcolsep{4pt}
			\renewcommand\arraystretch{1.0}
			\begin{tabular}{|c|r||c|c|c|c|c|c|c|}
				\hline\thickhline
				\rowcolor{mygray}
				
				Year &Method~~~~  &Pub. & Seg. Level  &Core Architecture 
& Flow &  Technical Feature & Training Dataset \\
				\hline
				\hline
				
				\multirow{3}{*}{\rotatebox{90}{2016}}
				& Clockwork~\cite{shelhamer2016clockwork} & ECCV & Semantic & FCN  
&  \checkmark& Faster Segmentation & Cityscapes~\cite{cordts2016cityscapes}/YouTube-Objects~\cite{DBLP:conf/cvpr/PrestLCSF12}  \\
				& FSO~\cite{kundu2016feature} & CVPR & Semantic& FCN + Dense CRF 
& \checkmark & Temporal Feature Aggregation & Cityscapes~\cite{cordts2016cityscapes}/CamVid~\cite{brostow2009semantic} \\
				& JFS~\cite{hur2016joint} & ECCV & Semantic& FCN 
& \checkmark & Temporal Feature Aggregation & KITTI MOTS~\cite{voigtlaender2019mots} \\
				
				\hline
				\multirow{5}{*}{\rotatebox{90}{2017}}
				& BANet~\cite{mahasseni2017budget} 					& CVPR & Semantic& FCN + LSTM 				
& 				& Keyframe Selection 		&  CamVid~\cite{brostow2009semantic}/KITTI\\
				& PEARL~\cite{jin2017video}                 & ICCV  & Semantic& FCN                
&	\checkmark		 & Flow-guided Feature Aggregation  			&  Cityscapes~\cite{cordts2016cityscapes}/CamVid~\cite{brostow2009semantic}\\
				& NetWarp~\cite{gadde2017semantic} & ICCV & Semantic& Siamese FCN  
& 	\checkmark		& Flow-guided Feature Aggregation 			&  Cityscapes~\cite{cordts2016cityscapes}/CamVid~\cite{brostow2009semantic}\\
				& DFF~\cite{zhu2017deep} 					& ICCV & Semantic& FCN  				
& 				& Flow-guided Feature Aggregation	&  Cityscapes~\cite{cordts2016cityscapes}\\
				& BBF~\cite{saleh2017bringing} & ICCV & Semantic& Two-Stream FCN 
& \checkmark & Weakly-Supervised Learning & Cityscapes~\cite{cordts2016cityscapes}/CamVid~\cite{brostow2009semantic} \\

				\hline
				\multirow{5}{*}{\rotatebox{90}{2018}}
				& GRFP~\cite{nilsson2018semantic} 		& CVPR & Semantic& FCN + GRU 		
&\checkmark & Temporal Feature Aggregation 		&Cityscapes~\cite{cordts2016cityscapes}/CamVid~\cite{brostow2009semantic} \\
				& LVS~\cite{li2018low}  							& CVPR & Semantic& FCN 				
& 					& Keyframe Selection 		&Cityscapes~\cite{cordts2016cityscapes}/CamVid~\cite{brostow2009semantic}\\
				& DVSN~\cite{xu2018dynamic}  				& CVPR & Semantic& FCN+RL 		
& \checkmark & Keyframe Selection  		&Cityscapes~\cite{cordts2016cityscapes}\\
				& EUVS~\cite{huang2018efficient} 			& ECCV & Semantic& Bayesian CNN 
& \checkmark & Flow-guided Feature Aggregation & CamVid~\cite{brostow2009semantic} \\
				& GCRF~\cite{chandra2018deep} 			& CVPR & Semantic& FCN+CRF 
 & \checkmark & Gaussian CRF & CamVid~\cite{brostow2009semantic} \\
				
				\hline
				\multirow{4}{*}{\rotatebox{90}{2019}}
				& Accel~\cite{jain2019accel} 				& CVPR 		& Semantic& FCN 				
&\checkmark   & Keyframe Selection			&  KITTI\\
				& SSeg~\cite{zhu2019improving} 				& CVPR 		& Semantic& FCN 				
&   & Weakly-Supervised Learning&  Cityscapes~\cite{cordts2016cityscapes}/CamVid~\cite{brostow2009semantic} \\
				& MOTS~\cite{voigtlaender2019mots} 				& CVPR 		& Instance& Mask R-CNN 				
&   & 	Tracking by Detection	&  KITTI MOTS~\cite{voigtlaender2019mots} /MOTSChallenge~\cite{voigtlaender2019mots}\\
				& MaskTrack R-CNN~\cite{yang2019video} & ICCV & Instance& Mask R-CNN 
& 						& Tracking by Detection				& YouTube-VIS~\cite{yang2019video}\\
				
				\hline
				\multirow{10}{*}{\rotatebox{90}{2020}}
				& EFC~\cite{ding2020every}					& AAAI 		& Semantic& FCN 				
& \checkmark	&Temporal Feature Aggregation 			& Cityscapes~\cite{cordts2016cityscapes}/CamVid~\cite{brostow2009semantic}\\
				& TDNet~\cite{hu2020temporally}			& CVPR 	& Semantic	& Memory Network 
&				& Attention-based Feature Aggregation 		&  Cityscapes~\cite{cordts2016cityscapes}/CamVid~\cite{brostow2009semantic}/NYUDv2~\cite{silberman2012indoor}\\
				& MaskProp~\cite{bertasius2020classifying}& CVPR & Instance & Mask R-CNN 
& & Instance Feature Propagation &  YouTube-VIS~\cite{yang2019video}\\
				& VPS~\cite{kim2020video} 		& CVPR & Panoptic& Mask R-CNN 
& & Spatio-Temporal Feature Alignment &VIPER-VPS~\cite{kim2020video}/Cityscapes-VPS~\cite{kim2020video} \\
				& MOTSNet~\cite{porzi2020learning} 		& CVPR & Instance& Mask R-CNN
& & Unsupervised Learning & KITTI MOTS~\cite{voigtlaender2019mots} /BDD100K~\cite{yu2020bdd100k} \\
				& MVAE~\cite{lin2020video} 		& CVPR & Instance& Mask R-CNN+VAE 
  & & Variational Inference  & KITTI MOTS~\cite{voigtlaender2019mots} /YouTube-VIS~\cite{yang2019video} \\
				& ETC~\cite{liu2020efficient}				& ECCV & Semantic& FCN + KD 
& \checkmark& Knowledge Distillation & Cityscapes~\cite{cordts2016cityscapes}/CamVid~\cite{brostow2009semantic}\\
				& Sipmask~\cite{cao2020sipmask}			& ECCV & Instance& FCOS 
  & & Single-Stage Segmentation & YouTube-VIS~\cite{yang2019video} \\
				& STEm-Seg~\cite{athar2020stem}		& ECCV & Instance& FCN 
&& Spatio-Temporal Embedding Learning & DAVIS$_{17}$~\cite{pont20172017}/YouTube-VIS~\cite{yang2019video}/KITTI-MOTS~\cite{voigtlaender2019mots}  \\
				& Naive-Student~\cite{chen2020naive}		& ECCV & Semantic& FCN+KD 
&& Semi-Supervised Learning & Cityscapes~\cite{cordts2016cityscapes} \\	
				\hline
				\multirow{11}{*}{\rotatebox{90}{2021}}
				& CompFeat~\cite{fu2021compfeat}& AAAI& Instance & Mask R-CNN 
& & Spatio-Temporal Feature Alignment & YouTube-VIS~\cite{yang2019video}\\
				& TraDeS~\cite{wu2021track} & CVPR & Instance & Siamese FCN 
& & Tracking by Detection & MOT/nuScenes/KITTI MOTS~\cite{voigtlaender2019mots} /YouTube-VIS~\cite{yang2019video}\\
				& SG-Net~\cite{liu2021sg} & CVPR & Instance & FCOS 
& & Single-Stage Segmentation & YouTube-VIS~\cite{yang2019video}\\
				& VisTR~\cite{wang2021end} & CVPR & Instance & Transformer 
& & Transformer-based Segmentation & YouTube-VIS~\cite{yang2019video}\\
				& SSDE~\cite{Hoyer_2021_CVPR} & CVPR & Semantic & FCN 
& & Semi-Supervised Learning & Cityscapes~\cite{cordts2016cityscapes} \\
				& SiamTrack~\cite{Woo_2021_CVPR} & CVPR & Panoptic& Siamese FCN 
& & Supervised Contrastive Learning & VIPER-VPS~\cite{kim2020video}/Cityscapes-VPS~\cite{kim2020video} \\
				& ViP-DeepLab~\cite{Qiao_2021_CVPR} & CVPR & Panoptic& FCN 
& & Depth-Aware Panoptic Segmentation & Cityscapes-VPS~\cite{kim2020video}  \\
				& fIRN~\cite{liu2021weakly} & CVPR & Instance& Mask R-CNN 
& \checkmark & Weakly-Supervised Learning & YouTube-VIS~\cite{yang2019video}/Cityscapes~\cite{cordts2016cityscapes} \\
				& SemiTrack~\cite{fu2021learning} & CVPR& Instance & SOLO 
&  & Semi-Supervised Learning & YouTube-VIS~\cite{yang2019video}/Cityscapes~\cite{cordts2016cityscapes} \\
& Propose-Reduce~\cite{lin2021video}& ICCV & Instance  &  Mask R-CNN  
& & Propose and Reduce & DAVIS$_{17}$~\cite{pont20172017}/YouTube-VIS~\cite{yang2019video}\\

	& {CrossVIS}~\cite{yang2021crossover} & {ICCV} & {Instance} & {FCN} & & {Dynamic Convolution} &  {YouTube-VIS~\cite{yang2019video}/OVIS~\cite{qi2021occluded}} \\

				\hline
			\end{tabular}
		}
	\end{threeparttable}
	\vspace{-8pt}
\end{table*}

	\vspace{-5pt}
\subsubsection{Language-guided$_{\!}$ Video$_{\!}$ Object$_{\!}$ Segmentation$_{\!}$ (LVOS)$_{\!\!\!}$}\label{sec:3.1.4}
LVOS is an emerging area, dating back to 2018~\cite{gavrilyuk2018actor,khoreva2018video}. Although there have already existed some efforts~\cite{gao2017tall} in the intersection of language and video understanding, none of them addresses pixel-level video-language reasoning. Most efforts in LVOS are made around the theme of efficient alignment between visual and linguistic modalities. According to the multi-modal information fusion strategy, existing models can be divided into three groups.

\noindent$\bullet$~\textbf{Dynamic Convolution-based Methods.} The first initiate was proposed in\!~\cite{gavrilyuk2018actor} that applies dynamic networks\!~\cite{li2017tracking} for visual-language relation modeling. Specifically, convolution filters, dynamically generated from linguistic query, are used to adaptively transform visual features into desired segments. In the same line of work, \cite{wang2020context,ningpolar} incorporate spatial context into filter generation. However, as indicated by~\cite{wang2019asymmetric}, {linguistic variation of input description may greatly impact sentence representation and subsequently make dynamic filters unstable, causing inaccurate segmentation.} For example, ``car in blue is parked on the grass'' and ``blue car standing on the grass'' have the same meaning but different generated filters, leading to poor performance.

\noindent$\bullet$~\textbf{Capsule Routing-based Methods.} In~\cite{mcintosh2020visual}, both video and textual inputs are encoded through capsules~\cite{hinton2018matrix}, which are considered effective in modeling visual/textual entities. Then,  dynamic routing is applied over the video and text capsules  for visual-textual information integration.

\noindent$\bullet$~\textbf{Attention-based Methods.} Neural attention technique is\\ also widely adopted in the filed of LVOS$_{\!}$~\cite{khoreva2018video,liu2021cross,seo2020urvos,ye2021referring,liang2022local}, for fully capturing global visual/textual context. \red{In~\cite{wang2019asymmetric}, vision-guided language attention and language-guided vision attention were developed to capture visual-textual correlations. In~\cite{hui2021collaborative}, two different attentions are learned to ground spatial and temporal relevant linguistic cues to static and dynamic visual embeddings, respectively. 
}

	\vspace{-15pt}
\subsection{Deep Learning-based VSS Models}\label{sec:3.2}

{Video semantic segmentation aims to group pixels with different semantics (\eg, category or instance membership), where different semantics result in different types of segmentation tasks, such as (instance-agnostic) video semantic segmentation (VSS, \S\ref{sec:vss}), video instance segmentation (VIS, \S\ref{sec:vis}) and video panoptic segmentation (VPS, \S\ref{sec:vps}).}

\subsubsection{$_{\!\!\!}$(Instance-agnostic)$_{\!}$  Video$_{\!}$ Semantic$_{\!}$ Segmentation$_{\!}$ (VSS)$_{\!\!\!\!\!\!\!}$}\label{sec:vss}
Extending the success of deep learning-based image semantic segmentation techniques to the video domain has become a research focus in computer vision recently. To achieve this, the most straightforward strategy is the na\"{i}ve application of an image semantic segmentation model in a frame-by-frame manner. But this strategy completely ignores temporal continuity and coherence cues  provided in videos. To make better use of temporal information, research efforts in this field are mainly made along two lines.

\noindent$\bullet$~\textbf{Efforts towards More Accurate Segmentation.} A major stream of methods exploits cross-frame relations to boost the prediction accuracy. They typically first apply the very same segmentation algorithms to each frame independently. Then they add extra modules on top, \eg, optical flow-guided feature aggregation~\cite{gadde2017semantic,jin2017video,huang2018efficient}, and sequential network based temporal information propagation~\cite{nilsson2018semantic}, to gather multi-frame context and get better results. For example, in some pioneer work~\cite{hur2016joint,kundu2016feature}, after performing static semantic segmentation for each frame individually, optical$_{\!}$ flow$_{\!}$~\cite{hur2016joint}$_{\!}$ or$_{\!}$ 3D$_{\!}$ CRF$_{\!}$~\cite{kundu2016feature}$_{\!}$ based$_{\!}$ post$_{\!}$ processing is applied for gaining temporally consistent segments. Later, \cite{chandra2018deep} jointly learns CNN-based per-frame segmentation and CRF-based spatio-temporal reasoning. In~\cite{gadde2017semantic}, features warped from previous frames with optical flow are combined with the current frame features for prediction. These methods require additional feature aggregation modules, which increase the computational costs during the inference. Recently, \cite{ding2020every} proposes to only incorporate flow-guided temporal consistency into the training phase, without bringing any extra inference cost. But its processing speed is still bounded to the adopted per-frame segmentation algorithms, as all
features must be recomputed at each frame. {For these methods, the utility in time-sensitive application areas, such as mobile and autonomous driving, is limited.}

\noindent$\bullet$~\textbf{Efforts towards Faster Segmentation.}  Yet another complementary line of work tries to leverage temporal information to accelerate computation. They approximate the expensive per-frame forward pass with cheaper alternatives, \ie, reusing the features in  neighbouring frames. In~\cite{shelhamer2016clockwork}, parts of segmentation networks are adaptively executed across frames, thus reducing the computation cost. Later methods use keyframes to avoid processing of each frame, and then propagate the outputs or the feature maps to other frames. For instance,~\cite{zhu2017deep} employs optical flow to warp the features between the keyframe and non-key frames. Adaptive keyframe selection is later exploited in~\cite{mahasseni2017budget,xu2018dynamic}, further enhanced by adaptive feature propagation~\cite{li2018low}. In~\cite{jain2019accel}, Jain \etal use a large, strong model to predict the keyframe and use a compact one in non-key frames. Keyframe-based methods have different computational
loads between keyframes and non-key frames, causing high maximum latency and unbalanced occupation of
computation resources that may decrease system efficiency~\cite{hu2020temporally}. Additionally,  the spatial misalignment of other frames with respect to the keyframes is challenging to compensate for and
often leads to different quantity results between keyframes and non-key frames. In~\cite{liu2020efficient}, a temporal consistency guided knowledge distillation technique is proposed to train a compact network, which is applied to all frames. In~\cite{hu2020temporally}, several weight-sharing sub-networks are distributed over sequential frames, whose extracted shallow features are composed for final segmentation. This trend of methods indeed speeds up inference, but still with the cost of reduced accuracy.

\noindent$\bullet$~\textbf{Semi-/Weakly-supervised based Methods.} Away from these main battlefields, some researchers made efforts to learn VSS under annotation efficient settings. In~\cite{saleh2017bringing}, classifier heatmaps are used to learn VSS from image tags~only. \cite{zhu2019improving,chen2020naive} use both labeled and unlabeled video frames to learn VSS. They propagate annotations from labeled frames to other unlabeled, neighboring frames~\cite{zhu2019improving}, or alternatively train teacher and student networks with groundtruth annotations and iteratively generated pseudo labels~\cite{chen2020naive}.

\subsubsection{{Video Instance Segmentation (VIS)}}\label{sec:vis}
 In 2019, Yang \etal extended image instance segmentation to the video domain~\cite{yang2019video}, which requires simultaneous detection, segmentation and tracking of instances in videos. This task is also known as \textit{multi-object tracking
and segmentation} (MOTS)~\cite{voigtlaender2019mots}. Based on the patterns of generating instance sequences, existing frameworks can be
roughly categorized into four paradigms: i) \textit{track-detect}, ii) \textit{clip-match}, iii) \textit{propose-reduce}, iv) \textit{segment-as-a-whole}. \red{Track-detect methods detect and segment instances for each individual frame, {followed by frame-by-frame instance tracking}~\cite{yang2019video,voigtlaender2019mots,fu2021compfeat,cao2020sipmask,lin2020video,liu2021sg,hu2020learning,porzi2020learning,wu2021track}. For example, in~\cite{yang2019video,voigtlaender2019mots,cui2021tf}, Mask R-CNN~\cite{he2017mask} is adapted for
VIS/MOTS by adding a tracking branch for cross-frame instance association. Alternatively, \cite{liu2021sg} models spatial attention to describe instances, tackling the task from a novel single-stage yet elegant perspective. Clip-match methods  divide an entire video into multiple  overlapped clips, and
perform VIS independently for each clip through mask propagation~\cite{bertasius2020classifying}  or spatial-temporal embedding~\cite{athar2020stem}. Final instance sequences are generated by  merging neighboring clips.} Both of the two paradigms need two independent steps to generate a complete sequence. They both generate multiple incomplete sequences (\ie, frames or clips) from a video, and merge (or complete) them by tracking/matching at the second stage. Intuitively, these paradigms are vulnerable to error accumulation in the process of merging sequences, especially when occlusion or fast motion exists. \red{To address these limitations, a propose-reduce paradigm is proposed in~\cite{lin2021video}. It first samples several key frames and obtains instance sequences by propagating the instance segmentation results from each key frame to the entire video. Then, the redundant sequence proposals of the same instances are removed.} {This paradigm not
only discards the step of merging incomplete sequences, but also achieves robust results considering multiple key frames. However, these three types of methods still need complex heuristic rules to associate instances and/or multiple steps to generate instance sequences.} \red{The segment-as-a-whole paradigm~\cite{wang2021end} is more elegant; it poses the task as a direct sequence prediction problem using Transformer~\cite{vaswani2017attention}.}

Almost all VIS models are built upon fully supervised learning, while \cite{liu2021weakly,fu2021learning} are the exceptions. Specifically, in~\cite{liu2021weakly}, motion and temporal consistency cues are leveraged to generate pseudo-labels from tag labeled videos for weakly supervised VIS learning. In \cite{fu2021learning}, a semi-supervised  embedding learning approach is proposed to learn VIS from pixel-wise annotated images and unlabeled videos.


\vspace{-4pt}
\subsubsection{Video Panoptic Segmentation (VPS)}\label{sec:vps}
Very recently, Kim \etal extended image panoptic segmentation to the video domain~\cite{kim2020video}, which aims at a holistic segmentation of all foreground instance tracklets and background regions, and assigning a semantic label to each video pixel. They adapt an image panoptic segmentation model~\cite{xiong2019upsnet} for VPS, by adding two modules for temporal feature fusion and cross-frame instance association, respectively. Later, temporal correspondence was explored in~\cite{Woo_2021_CVPR} through learning coarse segment-level and fine pixel-level matching. Qiao \etal~\cite{Qiao_2021_CVPR} propose to learn monocular depth estimation and video panoptic segmentation jointly.

\begin{figure}[t]
  \centering
      \includegraphics[width=0.99\linewidth]{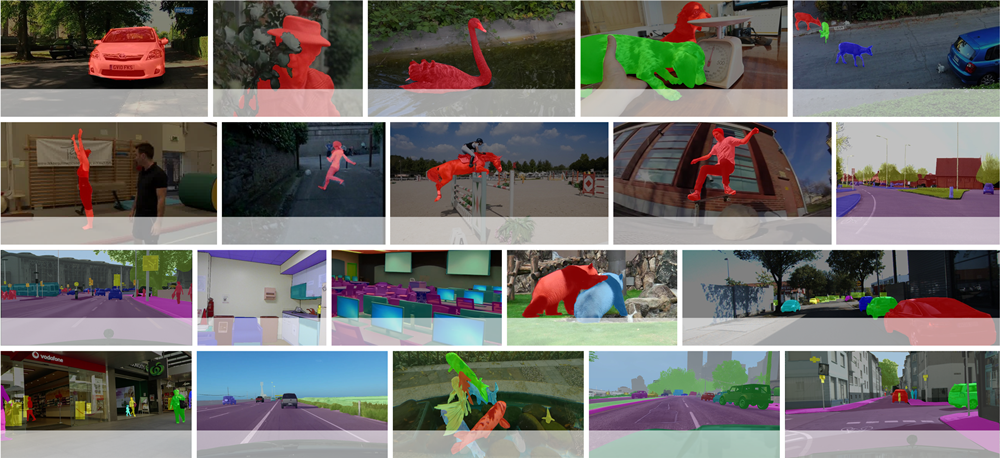}
       \put(-247,87.5){\tiny Youtube-Objects~\cite{DBLP:conf/cvpr/PrestLCSF12}}
      \put(-192,87.5){\tiny FBMS$_{59}\!$~\cite{DBLP:journals/pami/OchsMB14}}
      \put(-147.5,87.5){\tiny DAVIS$_{16}\!$~\cite{perazzi2016benchmark}}
      \put(-94,87.5){\tiny DAVIS$_{17}\!$~\cite{pont20172017}}
      \put(-47,87.5){\tiny YouTube-VOS~\cite{xu2018youtube}}
      \put(-245,55){\tiny A2D Sentence~\cite{gavrilyuk2018actor}}
      \put(-192,55){\tiny J-HMDB-S~\cite{gavrilyuk2018actor}}
      \put(-150,55){\tiny DAVIS$_{17}$-RVOS~\cite{khoreva2018video}}
      \put(-97,55){\tiny Refer-Youtube-VOS\!~\cite{seo2020urvos}}
      \put(-36,55){\tiny CamVid~\cite{brostow2009semantic}}
      \put(-244,30){\tiny {CityScapes}~\cite{cordts2016cityscapes}}
      \put(-201,30){\tiny NYUDv2\!~\cite{silberman2012indoor}}
      \put(-159,30){\tiny VSPW~\cite{miao2021vspw}}
      \put(-121,30){\tiny YouTube-VIS~\cite{yang2019video}}
      \put(-60,30){\tiny KITTI MOTS~\cite{voigtlaender2019mots}}
      \put(-250,1.9){\tiny {MOTSChallenge}\!~\cite{voigtlaender2019mots}}
      \put(-193.5,1.9){\tiny {BDD100K}~\cite{yu2020bdd100k}}
      \put(-140,1.9){\tiny {OVIS}~\cite{qi2021occluded}}
      \put(-98,1.9){\tiny {VIPER-VPS}~\cite{kim2020video} }
      \put(-50,1.9){\tiny Cityscapes-VPS~\cite{kim2020video}}
      	\vspace{-10pt}
\caption{{Example frames from twenty famous video segmentation benchmark datasets. The ground-truth segmentation annotation is overlaid.}}
     	\vspace{-10pt}
\label{fig:overviewdataset}
\end{figure}

\begin{table*}
\centering
\caption{\textbf{Statistics of representative video segmentation datasets.}
        See \S\ref{sec:4.1} and \S\ref{sec:4.2} for more detailed descriptions.}
\vspace{-7pt}
\begin{threeparttable}
\resizebox{0.99\textwidth}{!}{
\setlength\tabcolsep{8pt}
\renewcommand\arraystretch{1.0}
\begin{tabular}{|r||c|c|c|c|c|c|c|c|}
\hline\thickhline
\rowcolor{mygray}
Dataset~~~~~~~~~&Year &Pub. &\#Video &\#Train/Val/Test/Dev &Annotation &Purpose &\#Class &Synthetic\\
\hline
\hline
Youtube-Objects~\cite{DBLP:conf/cvpr/PrestLCSF12} &2012 &CVPR  &1,407
(126) &-/-/-/- &Object-level AVOS,  SVOS &Generic &10&\\
\rowcolor{mygray2}
FBMS$_{59}$~\!~~\cite{DBLP:journals/pami/OchsMB14} &2014 &PAMI &59 &29/30/-/- &Object-level AVOS,  SVOS &Generic&-&\\
DAVIS$_{16}$~\!~~\cite{perazzi2016benchmark} &2016 &CVPR &50 &30/20/-/- &Object-level AVOS,  SVOS &Generic&-&\\
\rowcolor{mygray2}
DAVIS$_{17}$~\cite{pont20172017} &2017 &- &150 &60/30/30/30 &Instance-level
AVOS, SVOS, IVOS &Generic&-&\\
YouTube-VOS~\cite{xu2018youtube} &2018 &- &4,519 &3,471/507/541/- &SVOS &Generic &94&\\
\rowcolor{mygray2}
A2D Sentence~\cite{gavrilyuk2018actor} &2018 &CVPR &3,782 &3,017/737/-/- &LVOS &Human-centric&-&\\
J-HMDB Sentence~\cite{gavrilyuk2018actor}&2018 &CVPR  &928 &-/-/-/- &LVOS &Human-centric&-&\\
\rowcolor{mygray2}
DAVIS$_{17}$-RVOS~\cite{khoreva2018video} &2018 &ACCV  &90 &60/30/-/- &LVOS &Generic&-&\\
Refer-Youtube-VOS~\cite{seo2020urvos} &2020 &ECCV  &3,975 &3,471/507/-/- &LVOS &Generic&-&\\
\rowcolor{mygray2}
CamVid~\cite{brostow2009semantic}  &2009 &PRL    &4  &(frame: 467/100/233/-) &VSS &Urban&11&\\
{CityScapes}~\cite{cordts2016cityscapes}   &2016  &CVPR    &5,000  &2,975/500/1,525 &VSS&Urban&19&\\
\rowcolor{mygray2}
NYUDv2~\cite{silberman2012indoor} &2012 &ECCV &518 &(frame: 795/654/-/-) &VSS &Indoor&40&\\
VSPW~\cite{miao2021vspw} &2021 &CVPR &3,536 &2,806/343/387/- &VSS &Generic&124&\\
\rowcolor{mygray2}
YouTube-VIS~\cite{yang2019video} &2019  &ICCV &3,859 &2,985/421/453/- &VIS &Generic&40&\\
KITTI MOTS~\cite{voigtlaender2019mots} &2019 &CVPR &21 &12/9/-/- &VIS &Urban &2&\\ 
\rowcolor{mygray2}
{MOTSChallenge}~\cite{voigtlaender2019mots} &2019 &CVPR &4 &-/-/-/- &VIS &Urban &1&\\
{BDD100K}~\cite{yu2020bdd100k} &2020 &ECCV &100,000 &7,000/1,000/2,000/- &VSS, VIS &Driving &40 (VSS), 8 (VIS)&\\ 
\rowcolor{mygray2}
{OVIS}~\cite{qi2021occluded} &2021 &- &901 & 607/140/154/-  &VIS &Generic&25&\\
{VIPER-VPS}~\cite{kim2020video} &2020 &CVPR &124 &(frame: 134K/50K/70K/-) &VPS &Urban&23 &\checkmark\\
\rowcolor{mygray2}
Cityscapes-VPS~\cite{kim2020video}&2020 &CVPR & 500 &400/100/-/- &VPS &Urban&19&\\
\hline
\end{tabular}
}
\end{threeparttable}
\label{table:dataset}
\vspace{-10pt}
\end{table*}

\vspace{-7pt}
\section{Video Segmentation Datasets} \label{sec:4}
Several datasets have been proposed for video segmentation over the past decades. {Fig.~\ref{fig:overviewdataset}
shows example frames from twenty commonly used datasets. We summarize
their essential features in Table~\ref{table:dataset} and give detailed review below.}

\vspace{-7pt}
\subsection{VOS Datasets}\label{sec:4.1}
\subsubsection{AVOS/SVOS/IVOS Datasets}
\noindent$\bullet$~\textbf{Youtube-Objects}  is a large dataset of $1,\!407$ videos collected from 155 web videos belonging to 10 object
categories (\eg, dog, cat, plane, \etc). VOS models typically test the generalization ability on a subset~\cite{jain2014supervoxel} having totally 126 shots with $20,\!647$ frames that provides coarse pixel-level fore-/background annotations on every 10$^{th}$ frames. 

\noindent$\bullet$~\textbf{FBMS$_{59}$}\!~\cite{DBLP:journals/pami/OchsMB14} consists of  59 video sequences with $13,\!860$ frames in total. However, only 720 frames are annotated for fore-/background separation. The dataset is split into 29 and 30 sequences for training and evaluation, respectively. 

\noindent$\bullet$~\textbf{DAVIS$_{16}$}\!~\cite{perazzi2016benchmark} has 50 videos (30 for \textit{train} set and 20 for \textit{val}
set) with $3,\!455$ frames in total. For each frame, in addition to high-quality fore-/background segmentation annotation, a set of attributes (\eg, deformation, occlusion, motion blur, \etc) are also provided to highlight the main challenges.

\noindent$\bullet$~\textbf{DAVIS$_{17}$}\!~\cite{pont20172017} contains 150 videos, \ie,  60/30/30/30 videos for \textit{train}/\textit{val}/\textit{test-dev}/\textit{test-challenge} sets. Its \textit{train} and \textit{val} sets are extended from the respective sets in DAVIS$_{16}$. There are 10,459 frames in total. DAVIS$_{17}$ provides instance-level annotations to support SVOS. Then, DAVIS$_{18}$ challenge\!~\cite{Caelles_arXiv_2018} provides scribble annotations to support IVOS. Moreover, as the original annotations of DAVIS$_{17}$ are biased towards the SVOS scenario, DAVIS$_{19}$ challenge\!~\cite{Caelles_arXiv_2019} re-annotates \textit{val} and \textit{test-dev} sets of DAVIS$_{17}$ to support AVOS. 

\noindent$\bullet$~\textbf{YouTube-VOS}\!~\cite{xu2018youtube} is a large-scale dataset, which is split into a  \textit{train} ($3,\!471$ videos), \textit{val} (507 videos), and \textit{test} (541 videos) set, in its newest 2019 version. Instance-level precise annotations are provided every five frames in a 30FPS frame rate. There are 94 object categories  (\eg, person, snake, \etc) in total, of which 26 are unseen in \textit{train} set.

\noindent \textbf{Remark.} Youtube-Objects, FBMS$_{59}$ and DAVIS$_{16}$ are used

\noindent for instance-agnostic AVOS and SVOS evaluation. DAVIS$_{17}$ is unique in comprehensive annotations for instance-level AVOS, SVOS as well as IVOS, but its scale is relatively small. YouTube-VOS is the largest one but only supports SVOS benchmarking. There also exist some other VOS datasets, such as SegTrack$_{V1}$~\cite{tsai2010} and SegTrack$_{V2}$~\cite{li2013video}, but they were less used recently, due to the limited scale and difficulty.

\subsubsection{LVOS Datasets}
\noindent$\bullet$~\textbf{A2D Sentence}\!~\cite{gavrilyuk2018actor} augments A2D~\cite{xu2015can} with phrases. It contains $3,\!782$ videos, with $8$ action classes performed by $7$ actors. In each video,  $3$ to $5$ frames are provided with segmentation masks. It contains $6,\!655$ sentences describing actors and their actions. The dataset is split into $3,\!017$/$737$ for \texttt{train}/\texttt{test}, and $28$ unlabeled videos are ignored~\cite{wang2019asymmetric}.

\noindent$\bullet$~\textbf{J-HMDB Sentence}\!~\cite{gavrilyuk2018actor} is built upon J-HMDB \cite{jhuang2013towards}. It  is comprised of $928$ short videos with $928$ corresponding sentences describing $21$ different action categories.

\noindent$\bullet$~\textbf{DAVIS$_{17}$-RVOS}\!~\cite{khoreva2018video} extends DAVIS$_{17}$ by collecting referring expressions for the annotated objects. 90 videos from \texttt{train} and \texttt{val} sets are annotated with more than 1,500 referring expressions. They provide two  types of annotations, which describe the highlighted
object: 1) based on the entire video (\ie, full-video expression) and 2) using only  the first frame of the video (\ie, first-frame expression).

\noindent$\bullet$~\textbf{Refer-Youtube-VOS}\!~\cite{seo2020urvos} includes  3,975 videos from {YouTube-VOS}\!~\cite{xu2018youtube},  with 27,899 language descriptions of target objects.
Similar to DAVIS$_{17}$-RVOS~\cite{khoreva2018video}, both full-video and first-frame expression annotations  are provided.

\noindent \textbf{Remark.} To date, {A2D Sentence} and {J-HMDB Sentence} are the main test-beds.  However, the phrases are not produced with the aim of reference, but description, and limited to only a few object categories corresponding to the dominant `actors' performing a salient `action'~\cite{seo2020urvos}. But newly introduced DAVIS$_{17}$-RVOS and Refer-Youtube-VOS show improved difficulties in both visual and linguistic modalities.

\vspace{-2pt}
\subsection{VSS Datasets}\label{sec:4.2}
\noindent$\bullet$~\textbf{CamVid}\!~\cite{brostow2009semantic} is composed of 4 urban scene videos with 11-class pixelwise annotations. Each video is annotated every 30 frames.  The
annotated frames are usually grouped into 467/100/233 for \texttt{train}/\texttt{val}/\texttt{test}~\cite{kundu2016feature}.

\noindent$\bullet$~\textbf{CityScapes}$_{\!~}$~\cite{cordts2016cityscapes}$_{\!~}$ is$_{\!~}$ a$_{\!~}$ large-scale$_{\!~}$ VSS$_{\!~}$ dataset$_{\!~}$ for$_{\!~}$ street

\noindent  views. It has
2,975/500/1,525 snippets for \texttt{train}/\texttt{val}/ \texttt{test}, captured at 17FPS. Each snippet contains 30 frames, and only the 20$^{th}$ frame is densely labelled with 19 semantic classes. 20,000 coarsely annotated frames are also provided.


\noindent$\bullet$~\textbf{NYUDv2}\!~\cite{silberman2012indoor} contains 518 indoor RGB-D videos with  high-quality ground-truths (every 10$^{th}$  video frame is labeled). There are 795 training frames and 654 testing frames being rectified and annotated with 40-class semantic labels.

\noindent$\bullet$~\textbf{VSPW}\!~\cite{miao2021vspw} is a recently  proposed large-scale VSS dataset. It addresses video scene parsing in the wild by considering diverse scenarios.  It consists of 3,536 videos, and provides  pixel-level annotations for 124 categories at 15FPS. The \texttt{train}/\texttt{val}/\texttt{test} sets contain 2,806/343/387 videos with 198,244/24,502/28,887 frames, respectively.

\noindent$\bullet$~\textbf{YouTube-VIS}\!~\cite{yang2019video} is built upon YouTube-VOS~\cite{xu2018youtube} with instance-level annotations. Its newest 2021 version has 3,859 videos (2,985/421/453 for \texttt{train}/\texttt{val}/\texttt{test}) with 40 semantic categories. It provides 232K high-quality annotations for 8,171 unique video instances.

\noindent$\bullet$~\textbf{KITTI MOTS}\!~\cite{voigtlaender2019mots} extends the 21 training sequences of KITTI tracking dataset~\cite{geiger2012we} with VIS annotations -- 12 for training and 9 for validation, respectively.  The dataset contains 8,008 frames with a resolution of $375\times1242$, 26,899 annotated cars and 11,420 annotated pedestrians.

\noindent$\bullet$~\textbf{MOTSChallenge}\!~\cite{voigtlaender2019mots} annotates 4 of 7 training sequences of MOTChallenge$_{2017}$~\cite{milan2016mot16}. It has 2,862 frames with 26,894 annotated pedestrians and presents many occlusion cases.

\noindent$\bullet$~\textbf{BDD100K}\!~\cite{yu2020bdd100k} is a large-scale dataset with 100K driving videos  (40 seconds and 30FPS each)  and supports  various tasks, including VSS and VIS. For VSS, 7,000/1,000/2,000 frames are densely labelled with 40 semantic classes for  \texttt{train}/\texttt{val}/\texttt{test}. For VIS, 90 videos with 8 semantic categories are annotated by 129K instance masks -- 60 training videos, 10 validation videos, and 20 testing videos.

\noindent$\bullet$~\textbf{OVIS}\!~\cite{qi2021occluded} is a new challenging
VIS dataset, where object occlusions usually occur. It has 901 videos and 296K high-quality instance masks for 25 semantic categories. It is split into 607 training, 140 validation and 154 test videos.

\noindent$\bullet$~\textbf{VIPER-VPS}\!~\cite{kim2020video} re-organizes VIPER~\cite{Richter_2017} into the video panoptic format. VIPER, extracted from the GTA-V game engine, has annotations of semantic and instance segmentations for 10 thing and 13 stuff classes on 254K frames of ego-centric driving scenes at $1080\!\times\!1920$  resolution.

\noindent$\bullet$~\textbf{Cityscapes-VPS}\!~\cite{kim2020video} is built upon CityScapes\!~\cite{cordts2016cityscapes}. Dense panoptic annotations for 8 thing and 11 stuff classes for 500 snippets in Cityscapes \texttt{val} set are provided every five frames and temporally consistent instance ids to the thing objects are also given, leading to 3000 annotated frames in total. These videos are split into 400/100 for \texttt{train}/\texttt{val}.

\noindent \textbf{Remark.} CamVid, CityScapes, NYUDv2, and VSPW are built for VSS benchmarking.  YouTube-VIS, OVIS, KITTI MOTS, and MOTSChallenge are VIS datasets, but the diversity of the last two are limited. BDD100K has both VSS and VIS  annotations. VIPER-VPS and Cityscapes-VPS are aware of VPS evaluation, but VIPER-VPS is a synthesized dataset.


\section{Performance Comparison} \label{sec:5}
{Next we tabulate the performance of previously discussed algorithms. For each of the reviewed fields, the most widely used dataset is selected for performance benchmarking. The performance scores are gathered from the original articles, unless specified. \red{For the running speed, we obtain the FPS for most methods by running their codes on a RTX 2080Ti GPU. For a small set of methods whose implementations are not well organized or publicly available, we directly borrow the values from the corresponding papers. Despite this, it is essential to remark the difficulty when comparing runtime. As different methods are with different code bases and levels of optimization, it is hard to make completely fair runtime comparison~\cite{marki2016bilateral,perazzi2016benchmark,garcia2018survey}; the values are only provided for reference.}


\subsection{Object-level AVOS Performance Benchmarking}
\subsubsection{Evaluation Metrics}
Presently, three metrics are frequently used~\cite{perazzi2016benchmark} to measure how object-level AVOS methods perform on this task:

\begin{table}[t]
	\centering
	\caption{{\textbf{Quantitative object-level AVOS results on
				DAVIS$_{16}$}~\cite{perazzi2016benchmark} \texttt{val}
			(\S\ref{sec:OAVOSeva}) in terms of region similarity $\mathcal{J}$, boundary
			accuracy $\mathcal{F}$ and time stability $\mathcal{T}$. We
			also report the recall and the decay performance over time
			for both $\mathcal{J}$ and $\mathcal{F}$. (FPS denotes \textit{frames per second}.  \red{$^\dagger$: FPS is borrowed from the original paper.} The three best scores are marked in \textcolor{red}{\textbf{red}}, \textcolor{blue}{\textbf{blue}}, and \textcolor{green}{\textbf{green}}, respectively.  These notes also apply to the other
			tables.)}}
\vspace{-5pt}
	\begin{threeparttable}
		\resizebox{0.48\textwidth}{!}{
			\setlength\tabcolsep{2.5pt}
			\renewcommand\arraystretch{1.0}
			\begin{tabular}{|r||ccc|ccc|c|c|}
				\hline\thickhline
				\rowcolor{mygray}
&\multicolumn{3}{c|}{{$\mathcal{J}$}} &\multicolumn{3}{c|}{{$\mathcal{F}$}}&$\mathcal{T}$ &\\
\rowcolor{mygray}
				\multirow{-2}{*}{{Method}}  &\textit{mean}$\uparrow$ &\textit{recall}$\uparrow$ &\textit{decay}$\downarrow$ &\textit{mean}$\uparrow$ &\textit{recall}$\uparrow$ &\textit{decay}$\downarrow$ &\textit{mean}$\downarrow$ & \multirow{-2}{*}{{FPS$\uparrow$}} \\ \hline \hline
				
				MuG~\cite{lu2020learning} & 58.0 & 65.3 & 2.0 & 51.5 & 53.2 & 2.1 & 30.1 & 2.5 \\
				 SFL~\cite{cheng2017segflow} &67.4 &81.4 &6.2 & 66.7& 77.1 &5.1 &28.2 & 3.3 \\
				 	
				 MotionGrouping~\cite{yang2021self} & 68.3 & - & - & 67.6 & -  & - & - & 83.3 \\
				  LVO~\cite{DBLP:conf/iccv/TokmakovAS17} &75.9 &89.1  &\textcolor{red}{\textbf{0.0}} &72.1  &83.4 &1.3 &26.5 & {13.5}\\
				  LMP~\cite{DBLP:conf/cvpr/TokmakovAS17} & 70.0 &85.0 &\textcolor{green}{\textbf{1.3}} & 65.9 & 79.2 &2.5&57.2 & {18.3}\\
			      FSEG~\cite{jain2017fusionseg} &70.7   &83.0 &1.5 &65.3 &73.8  &1.8  &32.8 & {7.2} \\
				PDB~\cite{Song_2018_ECCV}&77.2 &93.1 &\textcolor{blue}{\textbf{0.9}}  &74.5  &84.4  &\textcolor{red}{\textbf{-0.2}} &29.1 & 1.4 \\
				MOT~\cite{siam2018video}&77.2 &87.8 &5.0 &77.4 &84.4 &3.3 &27.9 & {1.0}\\
				LSMO~\cite{Tokmakov2019}&78.2 &91.1  &4.1 &75.9 &84.7 &3.5  &21.2 & 0.4  \\
				
				IST~\cite{Li_2018_CVPR} & 78.5 & - & - & 75.5 & - & - & - & - \\
				AGS~\cite{wang2019learning2}&79.7 &89.1 &1.9 &77.4 &85.8 &\textcolor{blue}{\textbf{0.0}} &26.7 & 1.7 \\
					
				MBN~\cite{Li_2018_ECCV1} & 80.4 & 93.2 & 4.8 & 78.5 & 88.6 & 4.4 & 27.8 & 1.0 \\
				COSNet~\cite{Lu_2019_CVPR}&80.5 &93.1  &4.4 &79.4 &89.5 &5.0 &\textcolor{blue}{\textbf{18.4}} & 2.2\\
				AGNN~\cite{wang2019zero} &81.3 &93.1 &\textcolor{red}{\textbf{0.0}} &79.7 &88.5 &5.1  &33.7&1.9\\
				
				MGA~\cite{Li_2019_ICCV} & 81.4 & - & - & 81.0 & - & - & - & 1.1 \\
				
				AnDiff~\cite{yang2019anchor} &81.7  &90.9 &2.2 &80.5 &85.1 &\textcolor{green}{\textbf{0.6}} &21.4& 2.8\\
				PyramidCSA~\cite{gu2020pyramid} &78.1  &90.1 &- &78.5  &88.2&-&-& $^\dagger$110 \\

				WCSNet~\cite{zhang2020unsupervised} & 82.2 & - & - & 80.7 & - & - & - & 25 \\
				
				MATNet~\cite{zhou2020motion} &82.4 &\textcolor{green}{\textbf{94.5}} &5.5 &80.7 &90.2 &4.5 &21.6&1.3\\
				EGMN~\cite{lu2020video} & 82.5 &94.3  &4.2 &81.2 &\textcolor{green}{\textbf{90.3}} &5.6 &\textcolor{green}{\textbf{19.8}}& {5.0} \\
				DFNet~\cite{zhen2020learning} &\textcolor{blue}{\textbf{83.4}} &-&-&\textcolor{green}{\textbf{81.8}}&-&- &\textcolor{red}{\textbf{15.9}} & $^\dagger$3.6\\
				F2Net~\cite{liu2021f2net} &\textcolor{green}{\textbf{83.1}} & \textcolor{blue}{\textbf{95.7}} &\textcolor{red}{\textbf{0.0}} &\textcolor{blue}{\textbf{84.4}} &\textcolor{blue}{\textbf{92.3}} &0.8 &20.9 & $^\dagger$10\\
				RTNet~\cite{Ren_2021_CVPR} &\textcolor{red}{\textbf{85.6}} &\textcolor{red}{\textbf{96.1}} & 0.5 &\textcolor{red}{\textbf{84.7}} &\textcolor{red}{\textbf{93.8}} &0.9 & 19.9 & {6.7}\\
				\hline
			\end{tabular}
		}
	\end{threeparttable}
	\label{table:OAVOS}
\vspace{-10pt}
\end{table}

\begin{table}[t]
	\centering
\renewcommand\thetable{8}
	\caption{{\textbf{Quantitative instance-level AVOS results on DAVIS$_{17}$}~\cite{pont20172017} \texttt{val} (\S\ref{sec:IAVOSeva}) in terms of region similarity $\mathcal{J}$ and boundary accuracy $\mathcal{F}$.}}
\vspace{-7pt}
	\begin{threeparttable}
		\resizebox{0.48\textwidth}{!}{
			\setlength\tabcolsep{2.5pt}
			\renewcommand\arraystretch{1.0}
			\begin{tabular}{|r||c|ccc|ccc|c|}
				\hline\thickhline
\rowcolor{mygray}
&$\mathcal{J}\!\&\mathcal{F}$ &\multicolumn{3}{c|}{{$\mathcal{J}$}} &\multicolumn{3}{c|}{{$\mathcal{F}$}}&\\
\rowcolor{mygray}
				\multirow{-2}{*}{{Method}} &\textit{mean}$\uparrow$ &\textit{mean}$\uparrow$ &\textit{recall}$\uparrow$ &\textit{decay}$\downarrow$ &\textit{mean}$\uparrow$ &\textit{recall}$\uparrow$ &\textit{decay}$\downarrow$ & \multirow{-2}{*}{{FPS$\uparrow$}} \\ \hline \hline
				 PDB~\cite{Song_2018_ECCV}    & 55.1 & 53.2 & 58.9 & 4.9 & 57.0 & 60.2 & \textcolor{green}{\textbf{6.8}} & 0.7\\
				
				 RVOS~\cite{ventura2019rvos} & 41.2 & 36.8 & 40.2 & \textcolor{green}{\textbf{0.5}} & 45.7 & 46.4 & 1.7 & 14.3\\
				  AGS~\cite{wang2019learning2}  & 57.5 & 55.5 & 61.6 & 7.0 & 59.5 & 62.8 & 9.0 & 1.1 \\
				 AGNN~\cite{lu2020zero} & 61.1 & 58.9 & 65.7 & 11.7 & 63.2 & 67.1 & 14.3 & {0.9} \\
				
				 STEm-Seg~\cite{athar2020stem}  & \textcolor{green}{\textbf{64.7}} & \textcolor{green}{\textbf{61.5}} & \textcolor{green}{\textbf{70.4}} & \textcolor{red}{\textbf{-4.0}} & \textcolor{blue}{\textbf{67.8}} & \textcolor{blue}{\textbf{75.5}} & \textcolor{blue}{\textbf{1.2}} & {9.3} \\
				  UnOVOST~\cite{luiten2020unovost} & \textcolor{red}{\textbf{67.9}} & \textcolor{red}{\textbf{66.4}} & \textcolor{red}{\textbf{76.4}} & \textcolor{blue}{\textbf{-0.2}} & \textcolor{red}{\textbf{69.3}} & \textcolor{red}{\textbf{76.9}} & \textcolor{red}{\textbf{0.0}} & $^\dagger$1.0 \\
				
				TODA~\cite{zhou2021target} & \textcolor{blue}{\textbf{65.0}} & \textcolor{blue}{\textbf{63.7}} & \textcolor{blue}{\textbf{71.9}} & 6.9 & \textcolor{green}{\textbf{66.2}} & \textcolor{green}{\textbf{73.1}} & 9.4 & 9.1 \\
				\hline
			\end{tabular}
		}
	\end{threeparttable}
	\label{table:IAVOS}
\vspace{-7pt}
\end{table}


\noindent$\bullet$~\textbf{Region Jaccard} $\mathcal{J}$ is calculated by the intersection-over-union (IoU) between the segmentation results ${\hat{Y}}\!\in\!\{0,1\}^{w\times h}$ \red{and the ground-truth ${Y}\!\in\!\{0,1\}^{w\times h}$:
$\mathcal{J} = {|\hat{Y} \cap  Y}|/|{\hat{Y}\cup Y|}$,}
which computes the number of pixels of the intersection between $\hat{Y}$ and ${Y}$, and divides it by the size of the union.

\noindent$\bullet$~\textbf{Boundary Accuracy}  $\mathcal{F}$ is the harmonic mean of the boundary precision $\text{P}_c$ and recall $\text{R}_c$. The value of $\mathcal{F}$ reflects how well the segment contours $c(\hat{Y})$ match the ground-truth contours $c(Y)$. Usually, the value of $\text{P}_c$ and $\text{R}_c$ can be computed via bipartite graph matching~\cite{martin2004learning}, then the boundary accuracy $\mathcal{F}$ \red{can be computed as: $\mathcal{F} = {2 \text{P}_c \text{R}_c}/({\text{P}_c + \text{R}_c})$}.

\noindent$\bullet$~\textbf{Temporal Stability}  $\mathcal{T}$ is informative of the stability of segments. It is computed as the pixel-level cost of matching two successive segmentation boundaries. The match is achieved by minimizing the shape context descriptor~\cite{belongie2002shape} distances between matched points while preserving the order in which the points are present in the boundary polygon. Note that $\mathcal{T}$
will compensate motion and small deformations, but not penalize inaccuracies of the contours~\cite{perazzi2016benchmark}.

\begin{figure*}
    \begin{minipage}{\textwidth}
        \begin{minipage}[t]{0.31\textwidth}
        \centering\small
  \makeatletter\def\@captype{table}
  \renewcommand\thetable{10}
  \caption{{\textbf{Quantitative IVOS results on DAVIS$_{17}$} \cite{pont20172017} \texttt{val} (\S\ref{sec:IVOSeva}) in terms of AUC and $\mathcal{J}$@60.}}\label{table:IVOS}
  \vspace{-5pt}
	\resizebox{1.\linewidth}{!}{
    \setlength\tabcolsep{15pt}
    \renewcommand\arraystretch{1.05}
    \begin{tabular}{|r||c|c|}
     \hline\thickhline
     \rowcolor{mygray}
     Method &AUC $\uparrow$ &$\mathcal{J}$@60 $\uparrow$\\ \hline \hline
     IVS~\cite{oh2019fast} &69.1 &73.4 \\
     MANet~\cite{miao2020memory} &74.9 &76.1 \\

     IVOS-W~\cite{Yin_2021_CVPR} & 74.1 & - \\
     ATNet~\cite{heo2020interactive} &\textcolor{green}{\textbf{77.1}} &\textcolor{green}{\textbf{79.0}}\\
     GIS~\cite{Heo_2021_CVPR} &\textcolor{blue}{\textbf{82.0}}&\textcolor{blue}{\textbf{82.9}}\\
     MiVOS~\cite{Cheng_2021_CVPRIVS}  &\textcolor{red}{\textbf{84.9}}  &\textcolor{red}{\textbf{85.4}}\\
     \hline
    \end{tabular}
   }
    \end{minipage}
\begin{minipage}[t]{0.69\textwidth}
\centering\small
\makeatletter\def\@captype{table}
\renewcommand\thetable{11}
\caption{{\textbf{Quantitative LVOS results on A2D Sentence}~\cite{gavrilyuk2018actor} \texttt{test}  (\S\ref{sec:LVOSeva}) \protect\\ in terms of {Precision@$K$}, {mAP} and {IoU}.}}	\label{table:a2dsota}
\vspace{-5pt}
	\resizebox{1.\linewidth}{!}{
		\setlength\tabcolsep{8pt}
		\renewcommand\arraystretch{1.05}
		\begin{tabular}{|r||ccccc|c|cc|c|}
			\hline\thickhline
			\rowcolor{mygray}
			   & \multicolumn{5}{c|}{{Overlap}} & {mAP$\uparrow$} & \multicolumn{2}{c|}{{IoU}} & \\\cline{2-6} \cline{8-9}
			\rowcolor{mygray}
			 \multirow{-2}{*}{{Method}} & P@0.5$\uparrow$ & P@0.6$\uparrow$ & P@0.7$\uparrow$ & P@0.8$\uparrow$ & P@0.9$\uparrow$ & 0.5:0.95 & \textit{overall}$\uparrow$ & \textit{mean}$\uparrow$ & \multirow{-2}{*}{{FPS$\uparrow$}} \\
			\hline \hline
			%

			 A2DS~\cite{gavrilyuk2018actor} & 50.0 & 37.6 & 23.1 & 9.4 & 0.4 & 21.5 & 55.1 & 42.6 & - \\
			 			
			 CMSANet~\cite{ye2021referring} & 46.7 & 38.5 & 27.9 & 13.6 & 1.7 & 25.3 & 61.8 & 43.2 & 6.5 \\
			AAN~\cite{wang2019asymmetric} & 55.7 & 45.9 & 31.9 & 16.0 & 2.0 & 27.4 & 60.1 & 49.0 & 8.6 \\
           VT-Capsule~\cite{mcintosh2020visual} & 52.6 & 45.0 & 34.5 & 20.7 & 3.6 & 30.3 & 56.8 & 46.0 & - \\
			CDNet~\cite{wang2020context} & \textcolor{green}{\textbf{60.7}} & \textcolor{green}{\textbf{52.5}} & \textcolor{green}{\textbf{40.5}} & \textcolor{green}{\textbf{23.5}} & \textcolor{green}{\textbf{4.5}} & \textcolor{green}{\textbf{33.3}} & \textcolor{green}{\textbf{62.3}} & \textcolor{blue}{\textbf{53.1}} & 7.2 \\
			 PolarRPE~\cite{ningpolar}& \textcolor{blue}{\textbf{63.4}} & \textcolor{blue}{\textbf{57.9}} & \textcolor{blue}{\textbf{48.3}} & \textcolor{blue}{\textbf{32.2}} & \textcolor{blue}{\textbf{8.3}} & \textcolor{blue}{\textbf{38.8}} & \textcolor{blue}{\textbf{66.1}} & \textcolor{green}{\textbf{52.9}} & 5.4 \\
            CST~\cite{hui2021collaborative} &\textcolor{red}{\textbf{65.4}} &\textcolor{red}{\textbf{58.9}} &\textcolor{red}{\textbf{49.7}} &\textcolor{red}{\textbf{33.3}} &\textcolor{red}{\textbf{9.1}} &\textcolor{red}{\textbf{39.9}} &\textcolor{red}{\textbf{66.2}} &\textcolor{red}{\textbf{56.1}} & {8.1} \\

			\hline
		\end{tabular}
	}
	\vspace{-14pt}
    \end{minipage}
    \end{minipage}
\end{figure*}

\vspace{-4pt}
\subsubsection{Results}\label{sec:OAVOSeva}
We select DAVIS$_{16}$\!~\cite{perazzi2016benchmark}, the most widely used dataset in AVOS, for performance benchmarking. Table~\ref{table:OAVOS} presents the results of those reviewed AVOS methods DAVIS$_{16}$ \texttt{val} set. The current best solution, RTNet~\cite{Ren_2021_CVPR}, reaches 85.6 region similarity $\mathcal{J}$, significantly outperforming the earlier deep learning-based methods, such as SFL~\cite{cheng2017segflow}, proposed in 2017.

\vspace{-5pt}
\subsection{Instance-level AVOS Performance Benchmarking}
\subsubsection{Evaluation Metrics}\label{sec:IAVOSm}
In instance-level AVOS setting, region Jaccard $\mathcal{J}$, boundary accuracy $\mathcal{F}$, and $\mathcal{J}\&\mathcal{F}$ -- the mean of $\mathcal{J}$ and $\mathcal{F}$ -- are used for evaluation~\cite{Caelles_arXiv_2019}. Each of the annotated object tracklets will be matched with one of predicted tracklets according to $\mathcal{J}\&\mathcal{F}$, using bipartite graph matching. For a certain criterion, the final score will be computed between each ground-truth object and its optimal assignment.

\subsubsection{Results}\label{sec:IAVOSeva}
Regarding instance-level AVOS, we take into account DAVIS$_{17}$~\cite{pont20172017} in which the vast majority of methods are evaluated. From Table~\ref{table:IAVOS} we can find that UnOVOST~\cite{luiten2020unovost} is the top scorer, with 67.9 $\mathcal{J}$  at the time of this writing.

\subsection{SVOS Performance Benchmarking}
\subsubsection{Evaluation Metrics}
Region Jaccard $\mathcal{J}$, boundary accuracy $\mathcal{F}$, and $\mathcal{J}\&\mathcal{F}$ are also widely adopted for SVOS performance evaluation~\cite{Caelles_arXiv_2018}.

\begin{table}[t]
  \centering
  	\vspace{-5pt}
  \renewcommand\thetable{9}
  \caption{{\textbf{Quantitative SVOS results on DAVIS$_{17}$}~\cite{pont20172017} \texttt{val} (\S\ref{sec:SVOSeva}) \protect\\ in terms of region similarity $\mathcal{J}$ and boundary accuracy $\mathcal{F}$.} }
  \vspace{-5pt}
  \begin{threeparttable}
   \resizebox{0.48\textwidth}{!}{
    \setlength\tabcolsep{1pt}
    \renewcommand\arraystretch{1.04}
    \begin{tabular}{|r||c|cc|c||r||c|cc|c|}
     \hline\thickhline
     \rowcolor{mygray}
     Method &\tabincell{c}{$\mathcal{J}\!\&\mathcal{F}$\\\textit{mean}$\uparrow$} & \tabincell{c}{$\mathcal{J}$\\\textit{mean}$\uparrow$} &\tabincell{c}{$\mathcal{F}$\\\textit{mean}$\uparrow$} & FPS$\uparrow$&Method &\tabincell{c}{$\mathcal{J}\!\&\mathcal{F}$\\\textit{mean}$\uparrow$} & \tabincell{c}{$\mathcal{J}$\\\textit{mean}$\uparrow$} &\tabincell{c}{$\mathcal{F}$\\\textit{mean}$\uparrow$} & FPS$\uparrow$\\ \hline \hline
     OnAVOS~\cite{voigtlaender2017online} &67.9 &64.5 &70.5 &0.08 &STM~\cite{Oh_2019_ICCV} &81.8 &79.2 &84.3 &6.3\\
     OSVOS~\cite{DBLP:conf/cvpr/CaellesMPLCG17} &60.3 &56.7 &63.9 &0.22 &e-OSVOS~\cite{meinhardt2020make} &77.2 &74.4 &80.0 &0.5\\
     CINM~\cite{Bao_2018_CVPR} &67.5 &64.5 &70.5 &$^\dagger$0.01 &AFB-URR~\cite{liang2020video} &74.6 &73.0  &76.1 &3.8\\
     FAVOS~\cite{cheng2018fast} &58.2 &54.6 &61.8 &0.56
      &Fasttan~\cite{huang2020fast} &75.9 &72.3 &79.4 &$^\dagger$7\\

      MAST~\cite{lai2020mast} & 65.5 & 63.3 & 67.6 & 5.1 & STM-Cycle~\cite{li2020delving}&  71.7 & 68.7 & 74.7 & 38 \\

      CRW~\cite{jabri2020space} & 67.6 & 64.5 & 70.6 & 7.3 &   QMA \cite{lin2021query} & 71.9 & - & - & 6.3 \\

     RGMP~\cite{wug2018fast} &66.7 &64.8 &68.6 &7.7 &Fasttmu~\cite{sun2020fast} &70.6 &69.1 &72.1 &9.7\\
     OSMN~\cite{yang2018efficient} &54.8 &52.5 &57.1 &7.7 &SAT~\cite{chen2020state} &72.3 &68.6 &76.0 &$^\dagger$39\\
     OSVOS-S~\cite{maninis2018video} &68.0 &64.7 &71.3 &0.22 &TVOS~\cite{zhang2020transductive} &72.3 &69.9 &74.7 &$^\dagger$37\\
     Videomatch~\cite{Hu_2018_ECCV} &61.4  &-&- &$^\dagger$0.38 &GCNet~\cite{li2020fast} &71.4 &69.3 &73.5 &$^\dagger$25\\
     Dyenet~\cite{Li_2018_ECCV} &69.1 &67.3 &71.0 &2.4 &KMN~\cite{seong2020kernelized} &76.0 &74.2 &77.8 &8.3\\
     MVOS~\cite{xiao2019online} &59.2  &56.3   &62.1 &1.5 &CFBI~\cite{yang2020collaborative} &81.9 &79.3 &84.5&2.2\\
     FEELVOS~\cite{voigtlaender2019feelvos} &71.5 &69.1 &74.0 &2.2 &LWL~\cite{bhat2020learning} &70.8 &68.2 &73.5 &15.6\\
     MHP-VOS~\cite{xu2019mhp} &75.3 &71.8 &78.8 &$^\dagger$0.01 &MSN~\cite{wu2020memory} &74.1 &71.4 &76.8&$\dagger$10\\
     AGSS~\cite{lin2019agss} &67.4 &64.9 &69.9 &$^\dagger$10 &EGMN~\cite{lu2020video} &\textcolor{blue}{\textbf{82.8}} &80.0 &85.2 &5.0 \\
     AGAME~\cite{johnander2019generative} &70.0 &67.2 &72.7 &$^\dagger$14 &SwiftNet~\cite{Wang_2021_SwiftNetCVPR} &81.1 &78.3 &83.9 &$^\dagger$25\\
     SiamMask~\cite{wang2019fast} &56.4 &64.3 &58.5 &$^\dagger$35 &G-FRTM~\cite{Park_2021_CVPR}&76.4& - &- &$^\dagger$18.2\\
     RVOS~\cite{ventura2019rvos} &60.6&57.5 &63.6 &0.56 &SST~\cite{duke2021sstvos} & 82.5 &79.9 &85.1&-\\
     RANet~\cite{wang2019ranet} &65.7 &63.2 &68.2 &$^\dagger$30 &GIEL~\cite{Ge_2021_CVPR} &\textcolor{green}{\textbf{82.7}} &\textcolor{green}{\textbf{80.2}} &\textcolor{green}{\textbf{85.3}} &6.7\\
     DMM-Net~\cite{Zeng_2019_ICCV} &70.7 &68.1 &73.3 &0.37 &LCM~\cite{Hu_2021_CVPR} &\textcolor{red}{\textbf{83.5}} &\textcolor{blue}{\textbf{80.5}} &\textcolor{red}{\textbf{86.5}} &8.5\\
     DTN~\cite{zhang2019fast} &67.4 &64.2 &70.6 &14.3 &RMNet~\cite{Xie_2021_CVPR} &\textcolor{red}{\textbf{83.5}}  &\textcolor{red}{\textbf{81.0}} &\textcolor{blue}{\textbf{86.0}} &$^\dagger$11.9\\
     \hline
    \end{tabular}
   }
  \end{threeparttable}
 \label{table:SVOS}
 \vspace{-11pt}
\end{table}

\vspace{-4pt}
\subsubsection{Results}\label{sec:SVOSeva}
DAVIS$_{17}$~\cite{pont20172017} is also one of the most important SVOS dataset.  Table~\ref{table:SVOS} shows the results of recent SVOS methods   on DAVIS$_{17}$ \texttt{val} set. In this case, all the top-leading solutions, such as EGMN~\cite{lu2020video}, LCM~\cite{Hu_2021_CVPR}, and RMNet~\cite{Xie_2021_CVPR}, are built upon the memory augmented architecture -- STM~\cite{Oh_2019_ICCV}.

	\vspace{-7pt}
\subsection{IVOS Performance Benchmarking}
	\vspace{-2pt}
\subsubsection{Evaluation Metrics}
	\vspace{-2pt}
Area under the curve (AUC) and Jaccard at 60 seconds ($\mathcal{J}$@60s) are two widely used IVOS evaluation criteria~\cite{Caelles_arXiv_2018}.

\noindent$\bullet$~\textbf{AUC} is designed to measure the overall accuracy of the evaluation. It is computed over the plot Time \textit{vs} Jaccard. Each sample in the plot is computed considering the average time and the average Jaccard for a certain interaction.

\noindent$\bullet$~\textbf{$\mathcal{J}$@60} measures the accuracy with a limited time budget (60 seconds). It is achieved by interpolating the Time \textit{vs} Jaccard plot at 60 seconds. This evaluates which quality an IVOS method can obtain in 60 seconds.

\vspace{-8pt}
\subsubsection{Results}\label{sec:IVOSeva}
DAVIS$_{17}$~\cite{pont20172017} is also widely used for IVOS performance benchmarking. Results summarized in Table~\ref{table:IVOS} show that the method proposed by Cheng \etal~\cite{Cheng_2021_CVPRIVS} is the top one.

\vspace{-11pt}
\subsection{LVOS Performance Benchmarking}
	\vspace{-2pt}
\subsubsection{Evaluation Metrics}
	\vspace{-2pt}
\red{As~\cite{gavrilyuk2018actor}, {overall IoU}, {mean IoU} and precision are adopted.}

\begin{table}[t]
	\centering
	\small
\renewcommand\thetable{12}
	\caption{{\textbf{Quantitative VSS results on Cityscapes}~\cite{cordts2016cityscapes} \texttt{val} (\S\ref{sec:VSSeva}) in terms of IoU$_{\text{class}}$ and IoU$_{\text{category}}$ (Max Latency: \textit{maximum per-frame time cost}).}}
	\vspace{-5pt}
\resizebox{0.45\textwidth}{!}{
		\setlength\tabcolsep{5pt}
		\renewcommand\arraystretch{1.0}
		\begin{tabular}{|r||cc|cc|}
			\hline\thickhline
			\rowcolor{mygray}
			{{Method}} &IoU$_{\text{class}}$$\uparrow$ &IoU$_{\text{category}}$ $\uparrow$ & {{FPS}$\uparrow$}  & {{Max Latency (ms)}$\downarrow$} \\
			\hline \hline
			Clockwork~\cite{shelhamer2016clockwork} & 66.4 & 88.6 & 6.4 &  198\\
			
			DFF~\cite{zhu2017deep} & 69.2 & 88.9 & 5.6  & 575\\
%
			PEARL~\cite{jin2017video} & 75.4 & 89.2 & 1.3 & 800 \\
			NetWarp~\cite{gadde2017semantic} & 80.5 & \textcolor{blue}{\textbf{91.0}} & -  & -\\	
			DVSN~\cite{xu2018dynamic} & 70.3 &  -& $^\dagger$19.8  & - \\
			LVS~\cite{li2018low} & 76.8 & 89.8 & 5.8 & 380 \\
			GRFP~\cite{nilsson2018semantic} & \textcolor{green}{\textbf{80.6}} & \textcolor{green}{\textbf{90.8}} & 3.9 & 255\\					
			 Accel~\cite{jain2019accel} & 75.5 & - & $^\dagger$1.1 & -\\
			 VPLR~\cite{zhu2019improving} & \textcolor{blue}{\textbf{81.4}} & - & $^\dagger${5.9} & -\\
			 TDNet~\cite{hu2020temporally} & 79.9 & 90.1 & 5.6 & 178\\
			 EFC~\cite{ding2020every} & \textcolor{red}{\textbf{83.5}} & \textcolor{red}{\textbf{92.2}} & $^\dagger${16.7} & - \\
			 Lukas~\cite{Hoyer_2021_CVPR} & 71.2 & - & $^\dagger${1.9}& -\\
			\hline
		\end{tabular}
	}
	
	\vspace{-10pt}
	\label{table:YVSSsota}
\end{table}

\begin{table*}[t]
	\centering
  \renewcommand\thetable{14}
	\small
\caption{{\textbf{Quantitative VPS results on Cityscapes-VPS}~\cite{kim2020video} (\S\ref{sec:VPSeva}) \texttt{test} in term of VPQ. Each cell shows
VPQ$^k$ / VPQ$^k$-Thing / VPQ$^k$-Stuff.}}
\vspace{-8pt}
	\resizebox{0.85\textwidth}{!}{
		\setlength\tabcolsep{9pt}
		\renewcommand\arraystretch{1.0}
		\begin{tabular}{|r||cccc|c|c|}
			\hline\thickhline
			\rowcolor{mygray}
			  & \multicolumn{4}{c|}{{Temporal window size}} & & \\\cline{2-5}
			\rowcolor{mygray}
			  \multirow{-2}{*}{{Method}} & $k=0$$\uparrow$ & $k=5$$\uparrow$ & $k=10$$\uparrow$ & $k=15$$\uparrow$ &\multirow{-2}{*}{VPQ$\uparrow$} & \multirow{-2}{*}{{FPS$\uparrow$}} \\
			\hline \hline
			 VPS~\cite{kim2020video} &64.2 / 59.0 / 67.7 &57.9 / 46.5 / 65.1 &54.8 / 41.1 / 63.4 &52.6 / 36.5 / 62.9 &57.4 / 45.8 / 64.8 &1.3 \\
			 SiamTrack~\cite{Woo_2021_CVPR} &63.8 / 59.4 / 66.6 &58.2 / 47.2 / 65.9 &56.0 / 43.2 / 64.4 &54.7 / 40.2 / 63.2 &57.8 / 47.5 / 65.0 &4.5\\
                ViP-DeepLab~\cite{Qiao_2021_CVPR}  &68.9 / 61.6 / 73.5 &62.9 / 51.0 / 70.5 &59.9 / 46.0 / 68.8 &58.2 / 42.1 / 68.4 &62.5 / 50.2 / 70.3  & {10.0}\\
\hline
		\end{tabular}
	}
	
	\vspace{-15pt}
	\label{table:vps}
\end{table*}

\noindent$\bullet$~\textbf{IoU}:  \textit{overall IoU} is computed as total intersection area of all test data over the total union area, while \textit{mean IoU} refers to average over IoU of each test sample.

\noindent\red{$\bullet$~\textbf{Precision}: Precision@$K$ is computed as the percentage of test samples whose IoU scores are higher than a threshold $K$. Precision at five thresholds ranging from 0.5 to 0.9 and\\ mean$_{\!}$ average$_{\!}$ precision$_{\!}$ (mAP)$_{\!}$ over$_{\!}$ 0.5:0.05:0.95$_{\!}$ are$_{\!}$ reported.}

\vspace{-6pt}
\subsubsection{Results}\label{sec:LVOSeva}

A2D Sentence~\cite{gavrilyuk2018actor} is arguably the most popular dataset~in

\noindent LVOS. Table~\ref{table:a2dsota} gives the results of six recent methods on A2D Sentence \texttt{test} set. It shows clear improvement trend from the first LVOS model~\cite{gavrilyuk2018actor} proposed in 2018, to recent complicated solution~\cite{hui2021collaborative}. {For runtime comparison, all the methods are tested on a video clip of 16 frames with resolution $512\!\times\!512$ and a textual sequence of {20} words.}

\vspace{-5pt}
\subsection{VSS Performance Benchmarking}
\subsubsection{Evaluation Metrics}
IoU metric is the most widely used metric in VSS. Moreover, in Cityscapes~\cite{cordts2016cityscapes} -- the gold-standard benchmark dataset in this field, two IoU scores, IoU$_{\text{category}}$ and IoU$_{\text{class}}$, defined over two semantic granularities, are reported.
Here, `category' refers to high-level semantic categories (\eg, vehicle, human), while `class' indicates more fine-grained semantic classes (\eg, car, bicycle, person, rider). In total,~\cite{cordts2016cityscapes} considers $19$ classes, which are further grouped into $8$ categories.

\vspace{-3pt}
\subsubsection{Results}\label{sec:VSSeva}
Table~\ref{table:YVSSsota} summarizes the results of eleven VSS approaches on Cityscapes~\cite{cordts2016cityscapes} \texttt{val} set.  
As seen, EFC~\cite{ding2020every} performs the best currently, with $83.5\%$ in terms of IoU$_{\text{class}}$.

\vspace{-3pt}
\subsection{VIS Performance Benchmarking}
\subsubsection{Evaluation Metrics}
As in~\cite{yang2019video}, precision and recall metrics are used for VIS performance evaluation. Precision at IoU thresholds 0.5 and 0.75, as well as mean average precision (mAP) over 0.50:0.05:0.95 are reported.  Recall@$N$ is defined as the maximum recall given $N$ segmented instances per video. These two metrics are first evaluated per category and then
averaged over the category set. The IoU metric is similar to region Jaccard $\mathcal{J}$ used in instance-level AVOS (\S\ref{sec:IAVOSm}).

\vspace{-3pt}
\subsubsection{Results}\label{sec:VISeva}
Table~\ref{table:YVISsota} gathers VIS results for on  YouTube-VIS~\cite{yang2019video} \texttt{val}  set, showing that Transformer-based architecture, \ie,  VisTR \cite{wang2021end}, and redundant sequence proposal based solution Propose-Reduce \cite{lin2021video}, greatly improve the state-of-the-art.

\vspace{-6pt}
\subsection{VPS Performance Benchmarking}
\subsubsection{Evaluation Metrics}
In~\cite{kim2020video}, the panoptic quality (PQ) metric used in image panoptic segmentation is modified as video panoptic quality (VPQ) to adapt to video panoptic segmentation.

\noindent\red{$\bullet$$_{\!}$~\textbf{VPQ}:$_{\!}$ Given$_{\!}$ a$_{\!}$ snippet$_{\!}$ $V^{t:t+k\!}$ with$_{\!}$ time$_{\!}$ window$_{\!}$ $k$,$_{\!}$ true$_{\!}$~po-$_{\!}$ sitive (TP) is defined by $\text{TP}\!=\!\{(u, \hat{u})_{\!}\!\in\!U_{\!}\!\times\!\hat{U}_{\!}\!:_{\!} \text{IoU}(u,\hat{u})\!>\!0.5\}$ where $U$ and $\hat{U}$ are
the set of the ground-truth and predicted tubes, respectively. False Positives (FP) and False Negatives (FN) are defined accordingly.  After accumulating TP$_c$, FP$_c$, and FN$_c$ on
all the clips with window size $k$ and class $c$,~we\\ define:$_{\!}$ $\text{VPQ}^k_{\!}\!=_{\!}\!\frac{1}{N_{\text{class}}}\!\sum_c\!\frac{\sum_{(u,\hat{u})\in\text{TP}_c}\!\text{IoU}(u,\hat{u})}{|\text{TP}_c|+\frac{1}{2}|\text{FP}_c|+\frac{1}{2}|\text{FN}_c|}$.$_{\!}$ When$_{\!}$ $k\!=\!1$,$_{\!}$ VPQ$^1$}

\noindent \red{is equivalent to PQ. For evaluation, $\text{VPQ}^k$ is reported over $k\!\in\!\{0,5,10,15\}$ and finally,  $\text{VPQ}\!=\!\frac{1}{4}\sum_{k\in\{0,5,10,15\}\!}\text{VPQ}^k$.}

\vspace{-5pt}
\subsubsection{Results}\label{sec:VPSeva}
\!Cityscapes-VPS\!~\cite{kim2020video}$_{\!}$ is$_{\!}$ chosen$_{\!}$ for$_{\!}$ testing$_{\!}$ VPS$_{\!}$ methods.$_{\!}$ As$_{\!}$ shown$_{\!}$ in$_{\!}$ Table\!~\ref{table:vps}, ViP-DeepLab\!~\cite{Qiao_2021_CVPR}$_{\!}$ is$_{\!}$ the$_{\!}$ top$_{\!}$ one.

\begin{table}[t]
	\centering  \renewcommand\thetable{13}
	\small
\caption{{\textbf{Quantitative VIS results on YouTube-VIS}~\cite{yang2019video}  the \texttt{val} (\S\ref{sec:VISeva}) \protect\\ in terms of {Precision@$K$}, {mAP}, Recall@$N$ and {IoU}.}}
\vspace{-8pt}
	\resizebox{0.47\textwidth}{!}{
		\setlength\tabcolsep{4.5pt}
		\renewcommand\arraystretch{1.0}
		\begin{tabular}{|r||cccc|c|c|}
			\hline\thickhline
			\rowcolor{mygray}
			 Method & P@0.5$\uparrow$ & P@0.75$\uparrow$ & R@1$\uparrow$ & R@10$\uparrow$ &\tabincell{c}{mAP$\uparrow$\\0.5:0.95} & FPS$\uparrow$ \\
			\hline \hline
			%
			
			fIRN~\cite{liu2021weakly} & 27.2 & 6.2 & 12.3 & 13.6 & 10.5 & 3 \\

			 MaskTrack R-CNN~\cite{yang2019video} &51.1 &32.6 &31.0 &35.5 &30.3 &20 \\
             Sipmask~\cite{cao2020sipmask} &53.0 &33.3 &33.5 &38.9 &32.5 &24\\
             STEm-Seg~\cite{athar2020stem} &55.8 &37.9 &34.4 &41.6 &34.6 & {9}\\

             CrossVIS~\cite{yang2021crossover} & 57.3 & 39.7 & 36.0 & 42.0 & 36.6 & 36 \\
             			
             SemiTrack~\cite{fu2021learning}  & 61.1 & 39.8 & 36.9 & 44.5 & 38.3 & 10 \\
             MaskProp~\cite{bertasius2020classifying}  &-  &\textcolor{blue}{\textbf{45.6}}  &-  &- &\textcolor{blue}{\textbf{42.5}}& $^\dagger${1}\\
             CompFeat~\cite{fu2021compfeat} &56.0 &38.6 &33.1 &40.3 &35.3  &$^\dagger${17}\\
             TraDeS~\cite{wu2021track} &52.6 &32.8 &29.1  &36.6 &32.6&26\\
             SG-Net~\cite{liu2021sg} &\textcolor{green}{\textbf{57.1}} &39.6 &\textcolor{green}{\textbf{35.9}} &\textcolor{green}{\textbf{43.0}} &36.3 &20\\
             VisTR~\cite{wang2021end}  &\textcolor{blue}{\textbf{64.0}} &\textcolor{green}{\textbf{45.0}} &\textcolor{blue}{\textbf{38.3}} &\textcolor{blue}{\textbf{44.9}} &\textcolor{green}{\textbf{40.1}}&58\\
             Propose-Reduce~\cite{lin2021video} &\textcolor{red}{\textbf{71.6}} &\textcolor{red}{\textbf{51.8}} &\textcolor{red}{\textbf{46.3}} &\textcolor{red}{\textbf{56.0}} &\textcolor{red}{\textbf{47.6}} &{2}\\
			\hline
		\end{tabular}
	}
	
	\vspace{-14pt}
	\label{table:YVISsota}
\end{table}

	\vspace{-10pt}
\subsection{Summary}
{From the results, we can draw several conclusions.
The most important of them is related to reproducibility. Across different video segmentation areas, many methods do not describe the setup for the experimentation or do not provide the source code for implementation. Some of them even do not release segmentation masks. Moreover,  different methods use various datasets and backbone models. These make fair comparison impossible and hurt reproducibility.}

\red{Another important fact discovered thanks to this study is the lack of information about execution time and memory use. Many methods particularly in the fields of AVOS, LVOS, and VPS, do not report execution time and almost no paper reports memory use. This void is due to the fact that many methods focus only on accuracy without any concern about running time efficiency or memory requirements.} However, in many application scenarios, such as mobile devices and self-driving cars,  computational power and memory are typically limited. As benchmark datasets and challenges serve as a main driven factor behind the fast evolution of segmentation techniques, we encourage organizers of future video segmentation datasets to give this kind of metrics its deserved importance in benchmarking.

Finally, performance on some extensively studied video segmentation datasets, such as DAVIS$_{16}$~\cite{perazzi2016benchmark} in AVOS, DAVIS$_{17}$~\cite{pont20172017} in SVOS, A2D Sentence~\cite{gavrilyuk2018actor} in LVOS, have nearly reached saturation. Though some new datasets are proposed recently and claim huge space for performance improvement, {the dataset collectors just gather more challenging samples, without necessarily figuring out which exact challenges have and have not been solved.} 



	\vspace{-13pt}
\section{Future Research Directions}\label{sec:6}
Based on the reviewed research, we list several future research directions {that we believe should be pursued.}

\noindent$\bullet$~\textbf{Long-Term Video Segmentation}: Long-term video segmentation is much closer to practical applications, such as video editing. However, as the sequences in existing datasets often span several seconds, the performance of VOS models over long video sequences (\eg, at the minute level) are still unexamined. Bringing VOS into the long-term setting will unlock new research lines, and put forward higher demand of the re-detection capability of VOS models.

\noindent$\bullet$~\textbf{Open World Video Segmentation}: Despite the obvious dynamic and open nature of the world, current VSS algorithms are typically developed in a closed-world paradigm, where all the object categories are known as a prior. These algorithms are often brittle once exposed to the realistic complexity of the open world, where they are unable to efficiently adapt and robustly generalize to unseen categories. For example, practical deployments of VSS systems in robotics, self-driving cars, and surveillance cannot afford to have complete knowledge on what classes to expect at inference time, while being trained in-house. This calls for smarter VSS systems, with a strong capability to identify unknown categories in their environments \cite{wang2021unidentified}.

{\noindent$\bullet$~\textbf{Cooperation across Different Video Segmentation Sub-fields}: VOS and VSS face many common challenges, \eg, object occlusion, deformation, and fast motion. Moreover, there are no precedents for modeling these tasks in a unified framework. Thus we call for closer collaboration across different video segmentation sub-fields.}

\noindent$\bullet$~\textbf{Annotation-Efficient Video Segmentation Solutions}: Though great advances have been achieved in various videos segmentation tasks, current top-leading algorithms are built on fully-supervised deep learning techniques, requiring a huge amount of annotated data. Though semi-supervised, weakly supervised and unsupervised alternatives were explored in some literature, annotation-efficient solutions receive far less attention and typically show weak performance, compared with the fully supervised ones. As the high temporal correlations in video data can provide additional cues for supervision, exploring existing annotation-efficient techniques in static semantic segmentation in the area of video segmentation is an appealing direction.


\noindent$\bullet$~\textbf{Adaptive Computation}: It is widely recognized that there exist high correlations among  video frames. Though such data redundancy and continuity are exploited to reduce the computation cost in VSS, almost all current video segmentation models are fixed feed-forward structures or work alternatively between heavy and light-weight modes. We expect more flexible segmentation model designs towards more efficient and adaptive computation~\cite{bengio2015conditional}, which allows network architecture change on-the-fly -- selectively activating part of the network in an input-dependent fashion.

\noindent$\bullet$~\textbf{Neural Architecture Search}: Video segmentation models are typically built upon hand-designed architectures, which may be suboptimal for capturing the nature of video data and limit the best possible performance. Using neural architecture search techniques to automate the design of video segmentation networks is a promising direction.

	\vspace{-8pt}
\section{CONCLUSION} \label{sec:7}
To$_{\!}$ our$_{\!}$ knowledge,$_{\!}$ this$_{\!}$ is$_{\!}$ the$_{\!}$ first$_{\!}$ survey$_{\!}$ to$_{\!}$ comprehensively\\  review recent progress in video segmentation. 
We~provided the reader with the necessary background knowledge and summarized more than 150 deep learning models according to various criteria, including task settings, technique contributions, and learning strategies. We also presented a structured survey of 20 widely used video segmentation datasets and benchmarking results on 7 most widely-used ones. We discussed the results and provided insight into the shape of future research directions and open problems in the field. In conclusion,  video segmentation has achieved notable progress thanks to the striking development of deep learning techniques, but several challenges still lie ahead.  
\ifCLASSOPTIONcaptionsoff
  \newpage
\fi



%
	\vspace{-4pt}
{\small
\bibliographystyle{IEEEtran}

\bibliography{egbib}
}

\vfill


\end{document}